\documentclass[10pt]{article} % For LaTeX2e
\usepackage{tmlr}
\setcitestyle{numbers,square,citesep={,}}
\usepackage{amsmath}
\usepackage{amssymb}
\usepackage{amsthm}
\usepackage{array}
\usepackage{multirow}
\usepackage{hyperref}
\usepackage{url}
\usepackage{graphicx}
\usepackage{capt-of}
\usepackage{adjustbox}
\usepackage{wrapfig}
\usepackage{placeins}
\usepackage{needspace}
\usepackage{booktabs}
\usepackage[table]{xcolor}
\usepackage{makecell}

\definecolor{gaintext}{HTML}{236B38}
\definecolor{backbonebg}{HTML}{EAF0F6}

\definecolor{routebg}{HTML}{DCEAF7}
\definecolor{spanbg}{HTML}{FFF0CC}
\definecolor{deniedbg}{HTML}{E2EFDA}

\definecolor{absection}{HTML}{F2E8D8} % Warm beige
\definecolor{aboverall}{HTML}{EAE4F3} % Soft lavender
\definecolor{abgain}{HTML}{DDF0EB}    % Pale teal

\definecolor{latencybg}{HTML}{EAE4F3}
\definecolor{transferbg}{HTML}{EAE4F3}
\definecolor{coveragebg}{HTML}{EAE4F3}
\definecolor{stressdrop}{HTML}{F8E4E4}
\definecolor{cigainbg}{HTML}{E5F2E8}
\definecolor{costbg}{HTML}{EAE4F3}
\definecolor{variantbg}{HTML}{EAE4F3}
\definecolor{detectorbg}{HTML}{FFF0CC}
\definecolor{controlbg}{HTML}{EAE4F3}
\definecolor{retainedbg}{HTML}{FFF0CC}
\definecolor{sparsebg}{HTML}{F2E8D8}

\allowdisplaybreaks[1]
\definecolor{existence}{HTML}{DCEAF7}
\definecolor{attribute}{HTML}{E2EFDA}
\definecolor{event}{HTML}{FCE4D6}
\definecolor{relation}{HTML}{E8DFF5}

\title{Negation Through Affirmation: Leveraging Positive Evidence in Vision-Language Models}

\author{\name Anonymous}

\begin{document}

\maketitle
\raggedbottom

\begin{abstract}
Frozen vision-language models often match concepts mentioned in a caption without reliably respecting their negation: a query for ``a kitchen with no toaster'' favors an image containing a toaster. Yet these models are substantially stronger at matching affirmative descriptions. We exploit this asymmetry with \emph{Negation-Aware Evidence Scoring} (NAES), which reformulates negated descriptions into affirmative queries and combines their evidence by caption structure. A lightweight text decomposer, NegSplit, extracts negated spans and the remaining affirmative context. For captions with an affirmative remainder, NAES matches that content and applies a presence-based correction for the denial. When no affirmative content remains, it queries the denied object's presence and reflects the evidence around a reference to score the negated description. This yields substantial accuracy gains: with LAION-2B CLIP, NAES achieves 82.4\% accuracy on COCO-MCQ and 93.0\% on VOC2007-MCQ, surpassing KTN by 2.9 and 0.9 percentage points; with OpenAI CLIP, it exceeds PeakPatch accuracy by 4.4 points on COCO-MCQ and 26.0 points on VOC2007-MCQ. NAES also improves negated-description retrieval on COCO and MSR-VTT. The strongest multiple-choice results use competing descriptions to center and align evidence; retrieval improvements persist without this conditioning. These findings show that separating affirmative matching from denied-object evidence can substantially improve negation-aware ranking. 
% Code is available at \url{https://anonymous.4open.science/r/naes-D72C}.
\end{abstract}

\begin{figure}[!ht]
\centering
\includegraphics[width=\linewidth]{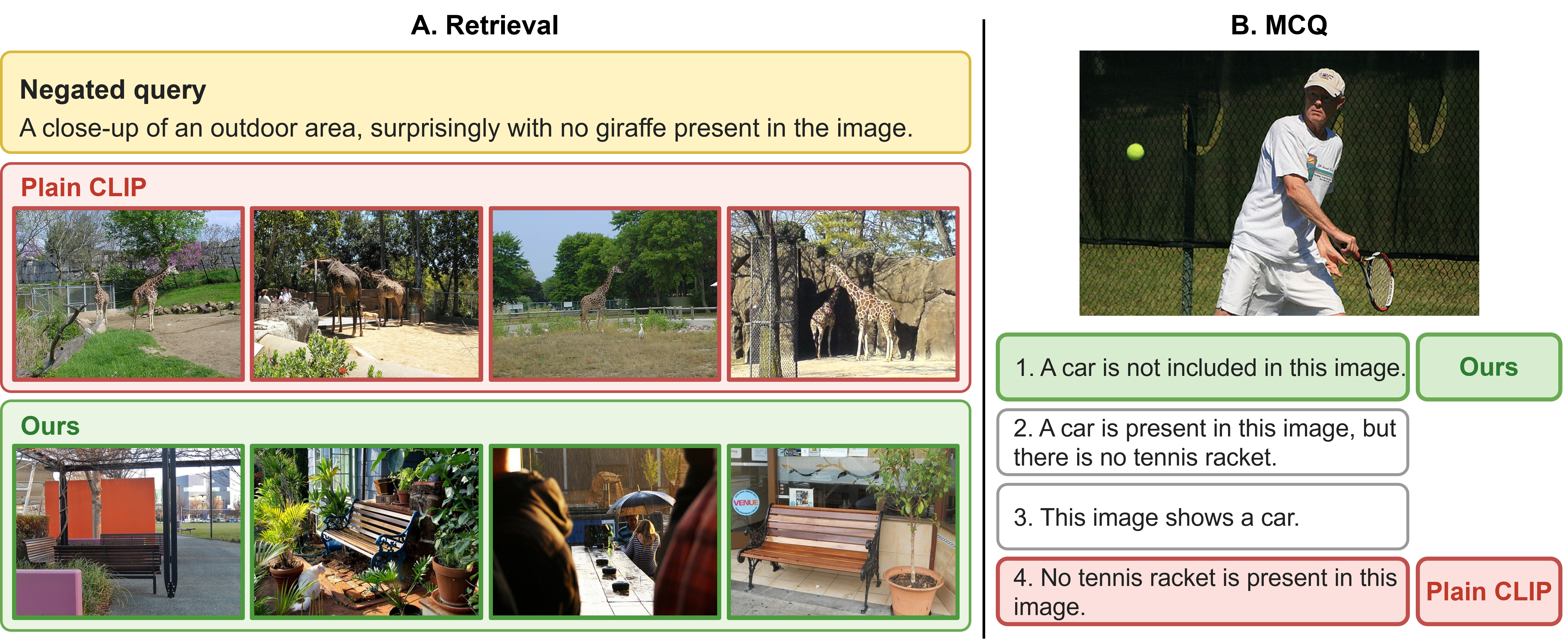}
\caption{Qualitative examples of negation-aware retrieval and MCQ selection. Left: for a query negating ``giraffe,'' plain CLIP's top-4 retrievals all contain a giraffe; ours excludes it at every rank. Right: for an MCQ question, plain CLIP picks ``No tennis racket is present in this image,'' while ours correctly selects ``A car is not included in this image.'' These illustrate the two scoring settings, not aggregate performance.}
\label{fig:teaser}
\end{figure}

\section{Introduction}
\label{sec:intro}

Contrastive vision-language models such as CLIP~\citep{radford2021clip} support retrieval and zero-shot recognition, but often struggle to distinguish what a description mentions from what it asserts. For example, consider a negation failure~\citep{alhamoud2025negbench}: ``a kitchen with no toaster'' may favor images containing a toaster, even though its presence contradicts the description. This behavior is consistent with earlier findings that vision-language models can exhibit bag-of-words-like matching, with limited sensitivity to word order and compositionality~\citep{yuksekgonul2023negclip}. Existing approaches address this failure through fine-tuning on negation-augmented data~\citep{yuksekgonul2023negclip,singh2025conclip,alhamoud2025negbench} or post-hoc correction of frozen models~\citep{lu2026peakpatch,ranjbar2025spacevlm,aggarwal2026seeingwhat,spacevlmdrc2026,kang2025dcsm,kim2026ktn}. We follow the latter direction, investigating how affirmative visual evidence can support negation-aware scoring without updating the pretrained encoders.

Our starting point is the asymmetry between affirmative and negated descriptions. Table~\ref{tab:gap} shows that native OpenAI CLIP achieves 69.1\% accuracy on COCO-MCQ when the correct answer is positive, compared with 39.3\% for hybrid and 6.8\% for negation-only answers. The last result is substantially below the 25\% chance level and is consistent with an \emph{affirmation bias} toward mentioned objects despite their negative polarity~\citep{alhamoud2025negbench}. Retrieval exhibits a corresponding gap: with the gallery and paired targets fixed, COCO R@1 decreases from 30.4\% for original captions to 25.0\% for their negated counterparts. These observations motivate exploiting the model's stronger affirmative matching capability when constructing scores for negated descriptions.

\begin{table}[!ht]
\centering
\small
\caption{Native OpenAI CLIP ViT-B/32 performance. MCQ accuracy (\%) is grouped by the correct answer's structure. Retrieval compares original captions (R) with their negated counterparts (R-Neg).}
\label{tab:gap}

\begin{tabular}{@{}lccc@{}}
\toprule
\multicolumn{4}{c}{MCQ accuracy (\%)} \\
\cmidrule(lr){2-4}
Dataset & Positive & Hybrid & Neg.-only \\
\midrule
COCO-MCQ~\citep{alhamoud2025negbench}
& 69.1 & 39.3 & 6.8 \\
VOC2007-MCQ~\citep{alhamoud2025negbench}
& 82.6 & 59.0 & 3.3 \\
\bottomrule
\end{tabular}
\hspace{2em}
\begin{tabular}{@{}lcccc@{}}
\toprule
\multicolumn{5}{c}{Retrieval (\%)} \\
\midrule
Dataset & R@1 & R-Neg@1 & R@5 & R-Neg@5 \\
\midrule
COCO~\citep{alhamoud2025negbench}
& 30.4 & 25.0 & 54.8 & 47.9 \\
MSR-VTT~\citep{alhamoud2025negbench}
& 29.0 & 24.7 & 51.8 & 47.6 \\
\bottomrule
\end{tabular}
\end{table}

The central challenge is to combine affirmative evidence according to the polarity of each claim. For ``a kitchen with no toaster,'' matching the affirmative remainder ``a kitchen'' establishes relevant scene content, but cannot distinguish kitchens with and without toasters. Evidence for a toaster must therefore contribute separately, with stronger presence evidence lowering the description's score. For ``no toaster,'' no affirmative remainder is available, so selection must rely on the evidence of the denied object. This distinction suggests complementary scoring rules for captions that retain affirmative content and those that contain only negation.

We introduce \emph{Negation-Aware Evidence Scoring} (NAES) to implement this principle. NegSplit, a text-only decomposer with approximately 2 million parameters, identifies caption structure, negated spans, and denied phrases. For hybrid captions, NAES constructs the affirmative remainder by removing the negated span and matches it to the image, with a gated correction based on evidence for the denied object. For negation-only captions, presence-score reflection assigns lower scores when evidence for a denied object is stronger. We consider a pool-independent rule with an external reference and a candidate-conditioned rule that uses categories mentioned across competing options to center evidence and align scores. Figure~\ref{fig:teaser} illustrates re-ranking in retrieval and multiple-choice selection. The scope of this evidence depends on what is negated. We distinguish between object existence, attributes or states, events, and relations. Object presence directly bears on an existence denial (e.g., ``a room with no skier''), but it does not resolve an event denial such as ``a skier not looking at the camera.'' This semantic distinction is separate from caption structure, which determines whether affirmative content remains after removing the negated span. 

Five of our six evaluation settings primarily target object-existence negation. The sixth, NegRefCOCOg~\citep{park2025knownobetter}, contains 440 expressions spanning all four semantic classes and examines the limits of object-based evidence. We evaluate NAES on two MCQ benchmarks: image and video retrieval, caption–foil discrimination, and referring-expression comprehension. With LAION-2B CLIP, candidate-conditioned accuracy reaches 82.4\% on COCO-MCQ and 93.0\% on VOC2007-MCQ, compared with CLIP accuracies of 24.8\% and 27.6\%. Retrieval improvements extend to SigLIP~\citep{zhai2023siglip}, SigLIP~2~\citep{tschannen2025siglip}, and larger CLIP encoders. Component ablations, paired confidence intervals, annotated extraction evaluation, and candidate-pool stress tests examine where these gains originate and how they depend on the available evidence. 

Our contributions are:
\begin{itemize}
    \item \textbf{A structure-dependent scoring framework:} We introduce NAES and the lightweight NegSplit decomposer, combining affirmative-remainder matching with presence-based correction and reflection while keeping the vision-language encoders frozen.
    \item \textbf{Evaluation and component analysis:} We evaluate six settings across multiple encoder families and quantify the contributions of decomposition, prompt rank fusion, and
    candidate-conditioned scoring through ablations with paired confidence intervals. A manually annotated subset examines extraction quality and its effect on downstream performance.
    \item \textbf{Analysis of semantic scope and robustness:} We distinguish caption structure from negation semantics and evaluate sensitivity to candidate-pool size and distractor composition, clarifying when affirmative evidence improves negation-aware selection.
\end{itemize}

\section{Related Work}
\label{sec:related}
\paragraph{Negation and compositionality in VLMs:} CLIP-based models often respond to mentioned concepts without reliably capturing their logical composition~\citep{yuksekgonul2023negclip}. Winoground~\citep{thrush2022winoground} tests word-order sensitivity, VL-Checklist~\citep{zhao2022vl} separates object, attribute, and relation understanding, and VALSE~\citep{parcalabescu2022valse} evaluates caption-foil discrimination across linguistic phenomena, including negation. NegBench~\citep{alhamoud2025negbench} extends this to negation in retrieval and multiple-choice tasks. A broader survey of negation in vision-language models is given in~\citep{pokhrel2026negationsurvey}. Because templates can admit nonvisual shortcuts, SugarCrepe~\citep{hsieh2023sugarcrepe} revises hackable benchmarks, motivating our own visual-dependence controls (Appendix~\ref{app:controls}). CREPE~\citep{ma2023crepe}, EqBen~\citep{wang2023equivariant}, and SPEC~\citep{peng2024synthesize} probe compositional generalization, equivariance, and fine-grained attributes. We instead distinguish the semantic type of a denied proposition from a caption's structural form.

\paragraph{Fine-tuning for negation:} The same NegBench work introduced above~\citep{alhamoud2025negbench} also fine-tunes CLIP variants on synthetic negation-augmented captions. CoN-CLIP~\citep{singh2025conclip} introduces a contrastive objective distinguishing affirmative from negated descriptions, while TNG-CLIP~\citep{cai2026tng} generates negated captions during training to strengthen sensitivity to negation. Structure-CLIP~\citep{huang2024structure} trains on scene-graph representations so that swapping two objects lowers image-caption similarity. TripletCLIP~\citep{patel2024tripletclip}, generative negative mining~\citep{sahin2024enhancing}, and cross-modal hard negative ranking~\citep{zhang2024contrasting} mine hard negatives for the same objective, including a medical-domain extension~\citep{vu2026negation}. Unlike these methods, our framework leverages pretrained encoders' affirmative recognition without retraining them.

\paragraph{Frozen-backbone correction:} Several approaches instead address negation without modifying the pretrained encoders. PeakPatch~\citep{lu2026peakpatch} corrects CLIP's scores with intermediate representations and a lightweight network, while DCSM~\citep{kang2025dcsm} replaces global similarity with patch-level scoring. SpaceVLM~\citep{ranjbar2025spacevlm} represents negated concepts as regions of embedding space; Aggarwal et al.~\citep{aggarwal2026seeingwhat} (denoted EC in our tables) reduce their influence through vector subtraction; and SpaceVLM-DRC~\citep{spacevlmdrc2026} adds context-aware repulsion. KTN~\citep{kim2026ktn} augments the frozen image embedding with a rejection dimension that penalizes the negated concept. Beyond negation, test-time adaptation methods correct frozen models using key-value caches~\citep{zhang2022tip}, entropy-minimizing prompts~\citep{shu2022test}, CLIP rewards~\citep{zhao2024test}, density-mode augmentation~\citep{zanella2024test}, or online clustering~\citep{hu2024bafta}. Our method scores negation-only captions post hoc with frozen-encoder similarities; only NegSplit is trained.

\paragraph{Scope extraction and visual evidence:} Frozen-backbone methods adjust similarity scores, while text-level methods identify negation cues and scope before scoring. The *SEM 2012 shared task~\citep{morante2012negation} frames negation as identifying a cue and its scope, and NegBERT~\citep{khandelwal2020negbert} applies contextual language representations to the same task. BioScope~\citep{szarvas2008bioscope}, the SFU Review Corpus~\citep{konstantinova2012review}, and CoNLL-2010~\citep{farkas2010conll} established scope annotation, later surveyed in~\citep{mahany2022negation}. Larger language models remain insensitive to negation~\citep{truong2023language}; Thunder-NUBench~\citep{so2026thunder} and a 200-cue-type study~\citep{tan2026not} further probe this gap. NegSplit jointly predicts the negated span, denied object, and caption structure, then removes the span to recover the affirmative remainder. For visual evidence, open-vocabulary detectors offer an alternative to CLIP similarity; we evaluate OWL-ViT~\citep{minderer2022owlvit} for VALSE-Existence only. We combine multiple prompt views via rank aggregation~\citep{dwork2001rankagg,cormack2009rrf}. We use a deterministic token-level rewrite instead of a learned model, and NAES's reflected reference acts as a scoring boundary, not a calibrated probability~\citep{guo2017calibration}.

\section{Problem Formulation}
\label{sec:formulation}

\paragraph{Negation in Vision-Language matching:}
We now formalize the matching objective underlying this challenge. Given an image $I$ and candidate descriptions $\mathcal{T}=\{t_1,\ldots,t_K\}$, we seek to rank the descriptions by their compatibility with the image. CLIP-like models~\citep{radford2021clip} score image–text pairs by cosine similarity between their embeddings:
\[
s_{\mathrm{VLM}}(I,t)
=
\frac{\langle f(I),g(t)\rangle}
{\|f(I)\|_2\,\|g(t)\|_2},
\]
where $f$ and $g$ are the image and text encoders. Negation poses a specific challenge to this matching objective: a description mentions a concept $c$ while denying a proposition $p$ about it. For example, ``a kitchen with no toaster'' mentions a toaster, but a toaster in the kitchen contradicts the description. A score driven by shared visual concepts can therefore favor an incompatible image. Successful matching requires distinguishing evidence for the mentioned concept $c$ from evidence for the proposition being denied.

\paragraph{Semantics of negation:} A negated description denies a proposition $p$. We distinguish four classes according to what $p$ expresses: object existence (``a room with no chair''), attribute or state (``a bird that is not fully visible''), event (``a skier not looking at the camera''), and relation (``a fruit not in a bowl'').
This taxonomy extends prior classifications~\citep{zhao2022vl,park2025knownobetter} by distinguishing event negation. Object absence addresses only existence negation. Occlusion and spatial relations require other evidence. We exclude numeric, quantifier, identity, and metalinguistic negation.

\paragraph{Caption structure:} Semantic class specifies what is denied. To score a caption, we classify its structure separately. Let $b(t)$ denote the negated span, including the negation expression and its scope, and let $a(t)$ denote the affirmative content remaining after removing $b(t)$. Following the three-way classification used in NegBench~\citep{alhamoud2025negbench}, we assign a structural label $z(t)\in \{\text{positive},\text{hybrid},\text{negation-only}\}$ to each candidate $t_k \in \mathcal{T}.$ A \emph{positive} caption contains no negation. A \emph{hybrid} caption contains both negated and affirmative content, leaving a meaningful remainder $a(t)$. A \emph{negation-only} caption leaves no meaningful affirmative content after its negated span is removed. We construct $a(t)$ by removing $b(t)$ from the original caption, rather than generating it with a separate model head.

\paragraph{Scoring objective:} The scoring route depends on $z(t)$: positive captions use their original text; hybrid captions use their affirmative remainder; and negation-only captions use affirmative evidence for the denied proposition, independent of semantic class. We distinguish pool-independent scoring, using a fixed external reference, from candidate-conditioned scoring, using the competing descriptions in $\mathcal{T}$. Section~\ref{sec:method} instantiates both, via $A_{v,k}$ for positive and hybrid captions and $R^{\mathrm{ind}}_k$/$R^{\mathrm{pool}}_{v,k}$ for negation-only captions.

More precisely, we aim to correct relative ranking rather than the calibrated probabilities of absence or the independent verification of negated propositions. We evaluate retrieval and candidate-selection performance, including ranking stability under changes to competing options.

\section{Method}
\label{sec:method}

\begin{figure}[!ht]
\centering
\includegraphics[width=\linewidth]{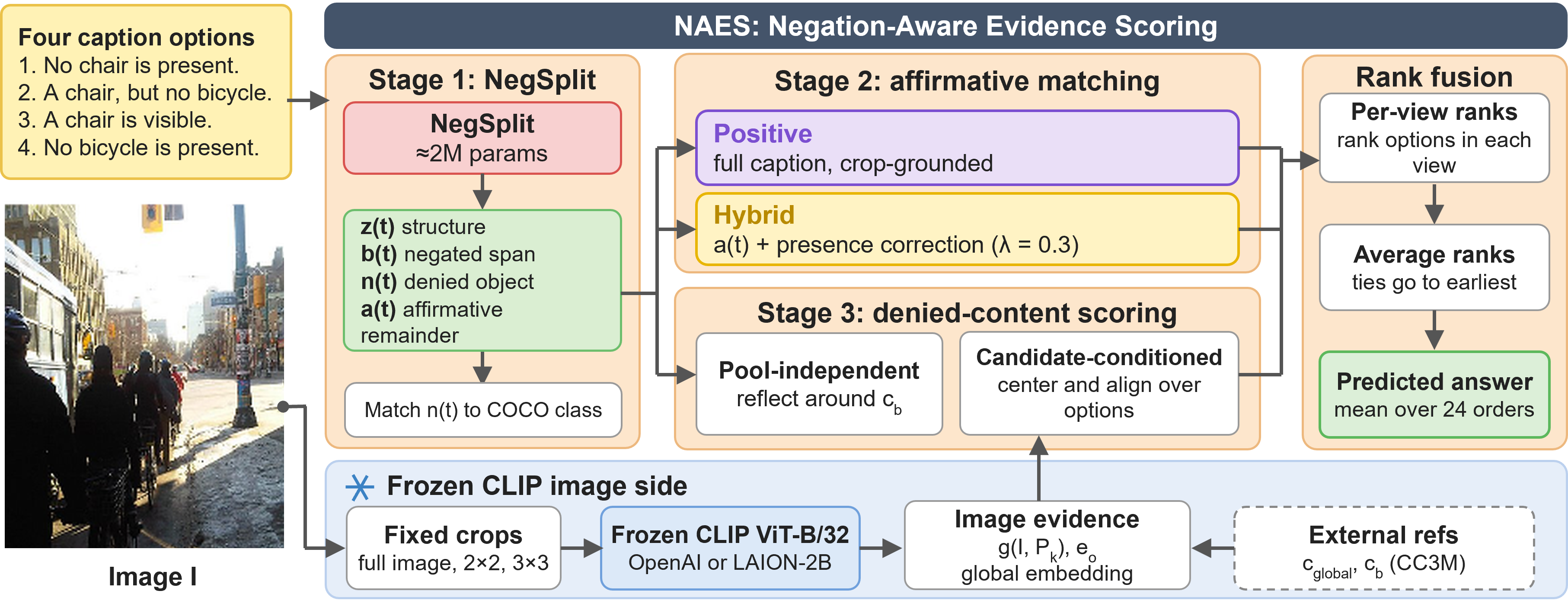}
\caption{Overview of NAES for MCQ. Stage 1: NegSplit extracts each option's negated span and denied object, leaving an affirmative remainder. Stage 2: frozen CLIP scores positive and hybrid remainders against crop-grounded, candidate-specific image vectors, with a fixed $\lambda=0.3$ presence correction for hybrid options. Stage 3: negation-only options are scored by reflecting denied-object evidence. Per-view ranks are fused.}
\label{fig:overview}
\end{figure}

We propose \emph{Negation-Aware Evidence Scoring} (NAES), which separates affirmative image-text matching from the scoring of denied content. The method has three components: a text decomposer that identifies the caption structure, an affirmative matching score for positive and hybrid captions, and a presence-based score for denied object mentions. These components define a structure-dependent scoring rule while keeping the CLIP encoders frozen. Figure~\ref{fig:overview} summarizes the MCQ pipeline.

\subsection{Caption decomposition}
\label{sec:decomposition}

We use NegSplit, a lightweight text-only encoder ($\approx$2M params), to extract the components defined in Section~\ref{sec:formulation}. Two token-level heads identify the negated span $b(t)$ and the denied phrase $n(t)$ within that span. Two sequence-level heads predict the structural label $z(t)$ and the semantic class.
The structural label determines the scoring rule; the semantic class is used only for stratified analysis (Appendix~\ref{app:negref}).

For hybrid captions, the affirmative remainder $a(t)$ is constructed by deleting $b(t)$ from the original text. It is not generated by a separate prediction head. This decomposition prevents the denied phrase from contributing directly to the affirmative matching score while retaining it for presence-based scoring. NegSplit is trained on synthetic text outside the evaluation benchmarks, without images or updates to the CLIP encoders. Appendix~\ref{app:setup} provides the architecture and training procedure.

\subsection{Affirmative matching}
\label{sec:rankfusion}

We set $x_k = t_k$ when $t_k$ is a positive caption, and $x_k = a(t_k)$ affirmative remainder, when $t_k$ is hybrid. We use two prompt templates, $\rho_1(x)=x$ and $\rho_2(x)=\text{``a photo of ''}+x$. For retrieval, the score against the gallery image $I_j$ is $A^{\mathrm{ret}}_{v,j} = \left\langle f_I(I_j),f_T(\rho_v(x))\right\rangle$, $v\in\{1,2\}$, where the encoder outputs are normalized.

For MCQ, we use a crop-aggregated image representation $g_{I,P_k}$ conditioned on $P_k$, the set of distinct object categories affirmed by candidate $k$: $A_{v,k}=\left\langle g_{I,P_k},f_T(\rho_v(x_k))\right\rangle$. The crop set contains the whole image and the cells of $2\times2$ and $3\times3$ grids. Attention over the affirmed categories $P_k$ determines the crop weights. Appendix~\ref{app:scoring} specifies this aggregation.

\paragraph{Presence-based correction:} For hybrid captions, affirmative matching leaves the denied content unaccounted for. We incorporate it through a global image-text query, $e^-_k=\left\langle f_I(I),f_T(\text{``a photo of a ''}+n(t_k))\right\rangle$ and adjust the affirmative score as
\begin{equation}
A'_{v,k}
=
A_{v,k}
+
\lambda(c_{\mathrm{global}}-e^-_k)
\operatorname{sigmoid}
\left(\frac{A_{v,k}-\mu_A}{\sigma_A}\right),
\label{eq:hybrid-correction}
\end{equation}
where $\mu_A,\sigma_A$ are the mean and standard deviation of native affirmative scores. The presence of evidence above the external reference $c_{\mathrm{global}}$ (defined below) penalizes the caption; evidence below it increases the score. The sigmoid gate limits this adjustment when the affirmative match is weak. We fix $\lambda=0.3$ using LAION-2B COCO retrieval and transfer it unchanged to the remaining settings. We estimate the reference and gate moments in Appendix~\ref{app:calibration}. The same correction applies to retrieval scores $A^{\mathrm{ret}}_{v,j}$, replacing $k$ with $j$ throughout; for retrieval, it is applied within each hybrid query's top-50 shortlist before full-gallery rank fusion (Appendix~\ref{app:scoring}). We refer to this adjustment as the hybrid correction.

\subsection{Scoring denied content}
\label{sec:ecpr}

For negation-only captions, no affirmative remainder is available. We instead ask whether the image supports the denied object. For example, ``no chair'' should receive a lower score when the image provides strong evidence for a chair. We apply two alternative scoring rules to localized evidence of object presence.

\begin{wrapfigure}[22]{r}{0.45\linewidth}
\centering
\includegraphics[width=\linewidth]{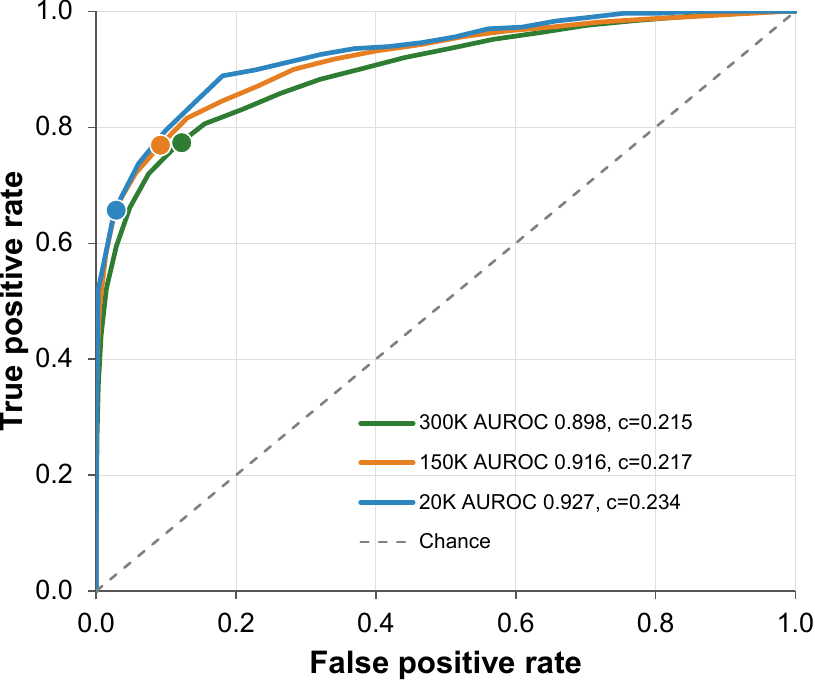}
\caption{Presence discrimination using whole-image CLIP similarities for OpenAI QuickGELU at three CC3M sample sizes. These curves assess the evidence used by the hybrid correction; localized reflection uses a separate presence fit.} 
\label{fig:cc3mroc}
\end{wrapfigure}

\paragraph{Finding presence evidence:} A deterministic matcher maps object mentions to the 80 COCO categories. For each resolved category, we compare its text embedding with the image crops and retain the strongest response. This allows a small object to contribute evidence even when it occupies little of the full image. Before selecting the strongest response, we subtract each crop's median category similarity to account for differences in its overall response level. An externally fitted affine map then aligns the resulting evidence, denoted $e_o(I)$, with the affirmative matching scale. Appendix~\ref{app:scoring} gives the full expression. This evidence concerns object presence. For example, detecting a bowl does not establish whether a fruit is inside it. The resulting scores, therefore, provide object-level evidence rather than verification of every denied proposition.

\paragraph{Establishing a presence reference:} To interpret the evidence, we estimate a reference that separates stronger from weaker presence responses. We fit logistic presence models on CC3M images using OWL-ViT-derived presence and absence pseudo-labels, without benchmark answer labels. The fitted model's zero-logit point defines the reference. The global hybrid correction and localized evidence use separate fits because their score scales differ. We denote their references by $c_{\mathrm{global}}$ and $c_b$, respectively, where $b$ identifies the backbone. The global fit uses 300,000 images. Figure~\ref{fig:cc3mroc} evaluates how well the global evidence distinguishes pseudo-labeled presence from absence across sample sizes. The ROC curves assess discrimination; the reference is obtained from the fitted presence model. Appendix~\ref{app:calibration} reports the fitting details and reference values.

\paragraph{Pool-independent reflection:}
We convert presence evidence into a denial score by reflecting it around $c_b$. Evidence above the reference produces a score below it, and vice versa:
\begin{equation}
R^{\mathrm{ind}}_k
=
\min_{o\in N_k}(2c_b-e_o)
=
2c_b-\max_{o\in N_k}e_o,
\label{eq:reflect}
\end{equation}
where $N_k$ contains the candidate's resolved denied categories. For ``no chair and no table,'' strong evidence for either object lowers the score. The minimum, therefore, treats multiple denials as a conjunction. This rule scores each candidate independently of the competing descriptions. The same reflection applies to retrieval, replacing $k$ with the gallery index $j$.

\paragraph{Candidate-conditioned scoring:} Alternatively, we interpret presence evidence relative to the categories mentioned across the candidate pool. For example, if chair evidence is stronger than table evidence, it favors ``a chair'' over ``a table'' and ``no table'' over ``no chair.''

With $P_k$ as defined above, let $N_k$ contain candidate $k$'s distinct denied categories, and let $U=\bigcup_k(P_k\cup N_k)$. We use their mean evidence as a shared center:
\begin{equation}
h=\frac{1}{|U|}\sum_{o\in U}e_o,
\qquad
C_k=
\frac{
\sum_{o\in P_k}(e_o-h)
+
\sum_{o\in N_k}(h-e_o)
}{
|P_k|+|N_k|
}.
\label{eq:ecpr-claims}
\end{equation}
Affirmed categories contribute positively above the center, while denied categories contribute positively below it. Averaging prevents candidates from gaining an advantage simply by mentioning more categories.

To compare these claim scores with affirmative matching scores, we standardize them across candidates and restore the mean and scale of the base scores:
\begin{equation}
R^{\mathrm{pool}}_{v,k}
=
\mu_v+\max(\sigma_v,\epsilon)
\frac{C_k-\bar C}{\sigma_C}.
\label{eq:ecpr-clip}
\end{equation}
Here $\bar C,\sigma_C$ are the pool moments of the claim scores, and $\mu_v,\sigma_v$ are those of the pre-replacement base scores in view $v$. Candidates without resolved categories retain their base scores and contribute $C_k=0$ to the pool statistics. When the pool provides insufficient categories or score variation, we use independent reflection when available; otherwise, we retain the base score. Appendix~\ref{app:scoring} specifies these fallback conditions.

\subsection{Final scoring and selection}
The caption structure determines which score is used: positive captions retain their affirmative matching scores, hybrid captions receive the presence-based corrections, and resolved negation-only captions receive either the independent or candidate-conditioned score. Other candidates retain their base scores. We combine the two prompt views by rank, avoiding direct averaging of their similarity scales. If $\pi_{v,k}$ is the descending rank of candidate $k$ under view $v$, we select $\widehat{k}=\arg\min_k\frac{\pi_{1,k}+\pi_{2,k}}{2}$. Ties are broken by the original candidate order. The independent rule preserves the reverse ordering of presence evidence; its reference positions these scores relative to positive and hybrid scores. The candidate-conditioned rule depends on the competing descriptions through its center and scale. We analyze this dependence in Appendix~\ref{app:pool}. Neither rule converts weak presence evidence into proof of absence: occlusion \& missed detection remain sources of error.

\section{Experimental Setup}
\label{sec:setup}

\paragraph{Benchmarks:} We evaluate six settings, five of which primarily target object-existence negation. COCO-MCQ and VOC2007-MCQ contain 5,914 and 5,031 four-choice questions, respectively~\citep{alhamoud2025negbench}. Text-to-image retrieval uses 5,000 COCO images and 25,014 captions~\citep{lin2014coco}; text-to-video retrieval uses 1,000 MSR-VTT videos~\citep{xu2016msrvtt}. VALSE-Existence contains 534 caption–foil pairs, of which 505 satisfy the annotator-agreement criterion~\citep{parcalabescu2022valse}. NegRefCOCOg contains 440 expressions with two candidate regions each, covering existence, attribute/state, event, and relation negation. We derive it from RefCOCOg~\citep{mao2016refcocog,park2025knownobetter}.

\paragraph{Models and comparisons:} We use frozen OpenAI CLIP ViT-B/32 with its original QuickGELU activation~\citep{radford2021clip} and LAION-2B CLIP ViT-B/32. Inference uses float32 precision with TF32 disabled unless stated otherwise. We compare NAES with matched, unmodified CLIP baselines and published methods. We qualify comparisons with published results where candidate access, evidence sources, or evaluation subsets differ. NegSplit uses either a transformer trained from scratch or a pruned BERT-tiny encoder~\citep{turc2019berttiny}, each with approximately 2 million parameters. Both are trained on 181,734 synthetic text examples with balanced structural classes and are evaluated without task-specific fine-tuning.

\begin{table}[!h]
\centering
\caption{Multiple-choice accuracy (\%) with CLIP ViT-B/32; ours average 24 option permutations. Pos., Neg., and Hyb.\ denote positive, negation-only, and hybrid answers. Best per backbone is bold; green shows point gains of conditioned NAES over matched CLIP. NAES uses the from-scratch decomposer with $\lambda=0.3$. Pool-independent scoring uses no cross-option information; candidate-conditioned scoring uses the candidate pool. The LAION-2B EC row is KTN's reimplementation.}
\label{tab:mcq}

\resizebox{\linewidth}{!}{%
\begin{tabular}{@{}c@{\hspace{6pt}}lc*{8}{r}@{}}
\toprule
& & & \multicolumn{4}{c}{\textbf{COCO-MCQ}}
      & \multicolumn{4}{c}{\textbf{VOC2007-MCQ}} \\
\cmidrule(lr){4-7}
\cmidrule(lr){8-11}
& \textbf{Method} & \textbf{Venue}
& Pos. & Neg. & Hyb. & \textbf{Avg}
& Pos. & Neg. & Hyb. & \textbf{Avg} \\
\midrule

\rowcolor{backbonebg}
\multicolumn{11}{@{}c@{}}{\textbf{OpenAI CLIP ViT-B/32}} \\
\addlinespace[2pt]

& CLIP~\citep{radford2021clip} & ICML'21
& 70.0 & 06.6 & 38.4 & 39.2
& 80.9 & 03.0 & 58.0 & 37.9 \\

& CLIP (matched) & --
& 69.1 & 06.8 & 39.3 & 39.3
& 82.6 & 03.3 & 59.0 & 38.7 \\

\midrule

\multirow{4}{*}{\rotatebox[origin=c]{90}{\textit{Finetune}}}
& NegCLIP~\citep{yuksekgonul2023negclip} & ICLR'23
& 49.2 & 13.9 & 16.3 & 26.8
& 70.5 & 04.6 & 42.3 & 30.2 \\

& CoN-CLIP~\citep{singh2025conclip} & WACV'25
& 15.6 & 32.9 & 25.3 & 24.4
& 24.8 & 23.2 & 56.7 & 38.2 \\

& CLIP+NegFull~\citep{alhamoud2025negbench} & CVPR'25
& 73.1 & 33.2 & 54.7 & 54.2
& 85.0 & 31.7 & 79.5 & 60.1 \\

& NegCLIP+NegFull~\citep{alhamoud2025negbench} & CVPR'25
& 81.0 & 25.9 & 60.1 & 56.5
& 81.0 & 21.1 & 83.7 & 58.2 \\

\midrule

\multirow{7}{*}{\rotatebox[origin=c]{90}{\textit{Post-hoc}}}
& DCSM~\citep{kang2025dcsm} & ICCV'25
& 71.2 & 06.6 & 68.0 & 48.6
& 68.5 & 05.4 & 73.1 & 49.0 \\

& SpaceVLM~\citep{ranjbar2025spacevlm} & arXiv'25
& 77.4 & 71.8 & 50.0 & 66.3
& 85.8 & 84.2 & 76.8 & 81.1 \\

& EC~\citep{aggarwal2026seeingwhat} & ICLR'26
& -- & -- & -- & 72.5
& -- & -- & -- & 78.6 \\

& PeakPatch~\citep{lu2026peakpatch} & ECCV'26
& \textbf{98.1} & 63.2 & 60.7 & 74.3
& \textbf{99.7} & 57.9 & 62.2 & 65.5 \\

\cmidrule(lr){2-11}

& NAES (independent) & \textbf{Ours}
& 76.8 & 68.1 & 65.8 & 70.3
& 88.7 & 71.3 & 87.9 & 81.0 \\

& NAES (conditioned) & \textbf{Ours}
& 76.6 & \textbf{84.5} & \textbf{75.5} & \textbf{78.7}
& 86.4 & \textbf{94.2} & \textbf{90.4} & \textbf{91.5} \\

& \multicolumn{2}{l}{\textcolor{gaintext}{$\Delta$ improvement}}
& \textcolor{gaintext}{+07.5}
& \textcolor{gaintext}{+77.7}
& \textcolor{gaintext}{+36.2}
& \textcolor{gaintext}{+39.4}
& \textcolor{gaintext}{+03.8}
& \textcolor{gaintext}{+90.9}
& \textcolor{gaintext}{+31.4}
& \textcolor{gaintext}{+52.8} \\

\midrule

\rowcolor{backbonebg}
\multicolumn{11}{@{}c@{}}{\textbf{LAION-2B CLIP ViT-B/32}} \\
\addlinespace[2pt]

& CLIP (matched) & --
& 44.5 & 14.0 & 14.9 & 24.8
& 66.3 & 05.4 & 37.0 & 27.6 \\

\midrule

\multirow{5}{*}{\rotatebox[origin=c]{90}{\textit{Post-hoc}}}
& EC (KTN reimpl.)~\citep{aggarwal2026seeingwhat} & ICLR'26
& -- & -- & -- & 70.0
& -- & -- & -- & 75.5 \\

& KTN~\citep{kim2026ktn} & arXiv'26
& -- & -- & -- & 79.5
& -- & -- & -- & 92.1 \\

\cmidrule(lr){2-11}

& NAES (independent) & \textbf{Ours}
& 78.2 & 73.2 & 72.2 & 74.6
& 90.5 & 69.7 & 89.7 & 81.4 \\

& NAES (conditioned) & \textbf{Ours}
& 80.8 & \textbf{87.8} & \textbf{79.1} & \textbf{82.4}
& 90.6 & \textbf{95.8} & \textbf{91.1} & \textbf{93.0} \\

& \multicolumn{2}{l}{\textcolor{gaintext}{$\Delta$ improvement}}
& \textcolor{gaintext}{+36.3}
& \textcolor{gaintext}{+73.8}
& \textcolor{gaintext}{+64.2}
& \textcolor{gaintext}{+57.6}
& \textcolor{gaintext}{+24.3}
& \textcolor{gaintext}{+90.4}
& \textcolor{gaintext}{+54.1}
& \textcolor{gaintext}{+65.4} \\

\bottomrule
\end{tabular}}
\end{table}

\paragraph{Reference estimation and parameter selection:} Presence references are fitted on CC3M~\citep{sharma2018cc3m} using OWL-ViT-derived pseudo-labels~\citep{minderer2022owlvit}, as described in Section~\ref{sec:method}. Separate backbone-specific references support the global hybrid correction and localized, pool-independent reflection. Candidate-conditioned scoring uses the competing options to construct its reference and falls back to the external reference when needed (Appendix~\ref{app:calibration}). We select the hybrid weight $\lambda=0.3$ using LAION-2B COCO retrieval performance and transfer it unchanged to the other settings and the OpenAI backbone. We inspected several transfer settings before fixing this choice; the selection is therefore informed by development.

\paragraph{Metrics:} MCQ accuracy is averaged over all 24 permutations of the four options and reported both overall and by the structural class of the correct answer. Retrieval uses R@1 and R@5 for original captions and R-Neg@1 and R-Neg@5 for their negated counterparts. For VALSE, a prediction is correct only when the caption scores strictly above its foil; ties count as errors. Where matched predictions are available, we report paired 95\% bootstrap confidence intervals from 10,000 resamples, clustered by source image or video.

\begin{table}[!h]
\centering
\caption{Retrieval recall (\%) with CLIP ViT-B/32. R@K and R-Neg@K use original and negated captions, respectively. Best negated-caption recalls within each backbone are bold; green values show percentage-point gains over matched CLIP. NAES uses pool-independent scoring, the from-scratch decomposer, and fixed $\lambda=0.3$; original-caption recall is unchanged. Our evaluation covers the full benchmarks. The LAION-2B EC result is KTN's reimplementation.}
\label{tab:retrieval}

\resizebox{\linewidth}{!}{%
\begin{tabular}{@{}c@{\hspace{6pt}}lc*{8}{r}@{}}
\toprule
& & & \multicolumn{4}{c}{\textbf{COCO}}
      & \multicolumn{4}{c}{\textbf{MSR-VTT}} \\
\cmidrule(lr){4-7}
\cmidrule(lr){8-11}
& \textbf{Method} & \textbf{Venue}
& R@1 & R-Neg@1 & R@5 & R-Neg@5
& R@1 & R-Neg@1 & R@5 & R-Neg@5 \\
\midrule

\rowcolor{backbonebg}
\multicolumn{11}{@{}c@{}}{\textbf{OpenAI CLIP ViT-B/32}} \\
\addlinespace[2pt]

& CLIP (reported)~\citep{radford2021clip} & ICML'21
& -- & 25.0 & -- & 47.9
& -- & 23.8 & -- & 45.9 \\

& CLIP (matched) & --
& 30.4 & 25.0 & 54.8 & 47.9
& 29.0 & 24.7 & 51.8 & 47.6 \\

\midrule

\multirow{2}{*}{%
\rotatebox[origin=c]{90}{\textit{FT}}}
& CoN-CLIP~\citep{singh2025conclip} & WACV'25
& -- & 25.7 & -- & 50.1
& -- & 23.3 & -- & 45.4 \\

& CLIP+NegFull~\citep{alhamoud2025negbench} & CVPR'25
& -- & 30.4 & -- & 55.0
& -- & 28.4 & -- & 51.6 \\

\midrule

\multirow{5}{*}{\rotatebox[origin=c]{90}{\textit{Post-hoc}}}
& SpaceVLM~\citep{ranjbar2025spacevlm} & arXiv'25
& -- & 29.9 & -- & 55.1
& -- & 26.1 & -- & 49.4 \\

& SpaceVLM-DRC~\citep{spacevlmdrc2026} & CVPRW'26
& -- & 30.2 & -- & 54.9
& -- & 28.9 & -- & 53.0 \\

& PeakPatch~\citep{lu2026peakpatch} & ECCV'26
& -- & 31.4 & -- & \textbf{56.9}
& -- & 29.1 & -- & 53.7 \\

\cmidrule(lr){2-11}

& NAES (independent) & \textbf{Ours}
& 30.4 & \textbf{33.3} & 54.8 & 56.4
& 29.0 & \textbf{30.6} & 51.8 & \textbf{54.0} \\

& \multicolumn{2}{l}{\textcolor{gaintext}{$\Delta$ improvement}}
& \textcolor{gaintext}{0.0}
& \textcolor{gaintext}{+8.3}
& \textcolor{gaintext}{0.0}
& \textcolor{gaintext}{+8.5}
& \textcolor{gaintext}{0.0}
& \textcolor{gaintext}{+5.9}
& \textcolor{gaintext}{0.0}
& \textcolor{gaintext}{+6.4} \\

\midrule

\rowcolor{backbonebg}
\multicolumn{11}{@{}c@{}}{\textbf{LAION-2B CLIP ViT-B/32}} \\
\addlinespace[2pt]

& CLIP (reported)~\citep{kim2026ktn} & arXiv'26
& -- & -- & 65.3 & 57.9
& -- & -- & 55.7 & 48.8 \\

& CLIP (matched) & --
& 38.8 & 31.9 & 64.8 & 57.3
& 30.6 & 26.4 & 54.5 & 48.2 \\

\midrule

\multirow{4}{*}{\rotatebox[origin=c]{90}{\textit{Post-hoc}}}
& EC (KTN reimpl.)~\citep{aggarwal2026seeingwhat} & ICLR'26
& -- & -- & 65.2 & 58.3
& -- & -- & 55.7 & 51.0 \\

& KTN~\citep{kim2026ktn} & arXiv'26
& -- & -- & 65.2 & 62.0
& -- & -- & 55.7 & 51.9 \\

\cmidrule(lr){2-11}

& NAES (independent) & \textbf{Ours}
& 38.8 & \textbf{39.9} & 64.8 & \textbf{64.7}
& 30.6 & \textbf{31.9} & 54.5 & \textbf{54.0} \\

& \multicolumn{2}{l}{\textcolor{gaintext}{$\Delta$ improvement}}
& \textcolor{gaintext}{0.0}
& \textcolor{gaintext}{+8.0}
& \textcolor{gaintext}{0.0}
& \textcolor{gaintext}{+7.4}
& \textcolor{gaintext}{0.0}
& \textcolor{gaintext}{+5.5}
& \textcolor{gaintext}{0.0}
& \textcolor{gaintext}{+5.8} \\

\bottomrule
\end{tabular}}
\end{table}

\section{Results \& Discussions}
\label{sec:results}

We evaluate multiple-choice selection and retrieval, then conduct component ablations, qualitative analysis, encoder transfer, and inference cost analysis. CLIP remains frozen throughout. Additional analyses appear in Appendices~\ref{app:support} and~\ref{app:qualify}.

\subsection{Main results}
\label{sec:main-results}

\paragraph{MCQ:} NAES substantially improves selection without updating the vision-language encoders (Table~\ref{tab:mcq}). On COCO-MCQ, accuracy increases from 39.3\% to 78.7\% with OpenAI CLIP and from 24.8\% to 82.4\% with LAION-2B; VOC2007-MCQ accuracy reaches 91.5\% and 93.0\%, respectively. The strongest improvements concern negation-only answers: OpenAI accuracy rises from 6.8\% to 84.5\% on COCO and from 3.3\% to 94.2\% on VOC. For these captions, affirmative decomposition alone provides no matching text. Reflection (Eq.~\ref{eq:reflect}) instead provides evidence for a denied object count against the description.

Candidate conditioning contributes beyond this reversal. On LAION-2B COCO-MCQ, independent scoring achieves 74.6\%, compared with 82.4\% under candidate conditioning. The shared reference and score mapping (Eqs.~\ref{eq:ecpr-claims}--\ref{eq:ecpr-clip}) allow denied-object evidence to compete with affirmative scores on a common scale. Positive-answer accuracy also remains above matched CLIP across both backbones and datasets, indicating that the negation gains do not require sacrificing affirmative selection. The COCO result exceeds published KTN accuracy by 2.9 points, although this is not a matched comparison. These findings support improved relative scoring; the additional candidate-conditioned gain depends on access to competing descriptions.

\begin{wraptable}{r}{0.45\textwidth}

\centering
\vspace{-2em}
\caption{Decomposer performance (\%): Structure uses 400 examples; spans use 344 valid annotations each, reported as F1/EM. Blue highlights structural classification, yellow negated-span EM, and green denied-object extraction. Bold indicates metric-wise best.}
\label{tab:decomposer-quality}

\scriptsize
\setlength{\tabcolsep}{2.5pt}
\renewcommand{\arraystretch}{1.18}

\begin{tabular}{@{}lcccc@{}}
\toprule
& \multicolumn{2}{c}{\textbf{Structure}}
& \multicolumn{2}{c}{\textbf{Spans: F1/EM}} \\
\cmidrule(lr){2-3}
\cmidrule(lr){4-5}
\textbf{System} & Acc. & MF1 & Negated & Denied \\
\midrule
Lexical parser
& 91.0 & 86.6 & 82.7 / 30.8 & 66.1 / 37.8 \\

HMM
& 87.0 & 70.4 & 74.3 / 28.8 & 49.8 / 38.4 \\

spaCy
& 83.8 & 80.7 & 85.9 / 44.5
& \cellcolor{deniedbg}\textbf{76.5} / \textbf{54.9} \\

DeBERTa cascade
& 82.3 & 77.1 & 78.6 / 47.1 & -- \\
\midrule
NegSplit (scratch)
& 95.3 & 92.5
& \cellcolor{spanbg}90.4 / \textbf{72.4}
& 71.5 / 45.6 \\

NegSplit (BERT)
& \cellcolor{routebg}\textbf{96.3}
& \cellcolor{routebg}\textbf{93.9}
& \textbf{92.2} / 70.9 & 73.3 / 45.9 \\
\bottomrule
\end{tabular}
\vspace{-2em}
\end{wraptable}

\paragraph{Retrieval:} The retrieval results show that the reformulation also helps without competing answer descriptions (Table~\ref{tab:retrieval}). On OpenAI COCO, R-Neg@1 increases from 25.0\% to 33.3\% and R-Neg@5 from 47.9\% to 56.4\%. On MSR-VTT, R-Neg@1 improves from 24.7\% to 30.6\%. Affirmative-remainder matching reduces the influence of denied concepts on the initial ranking; the correction in Eq.~\ref{eq:hybrid-correction} subsequently penalizes shortlisted images that support the denied content. The benefit also transfers to LAION-2B, where COCO R-Neg@5 rises from 57.3\% to 64.7\%. This exceeds KTN's reported 62.0\%, while OpenAI COCO R-Neg@5 remains slightly below PeakPatch (56.4\% versus 56.9\%). These comparisons use published results rather than matched reruns. Original-caption recall remains unchanged by construction.

\paragraph{Decomposer variants:} NegSplit's strongest extraction results align with the decisions needed for affirmative-remainder matching (Table~\ref{tab:decomposer-quality}). The scratch and BERT-tiny variants achieve structural classification accuracies of 95.3\% and 96.3\%, compared with 91.0\% for the lexical parser, and negated-span F1 scores of 90.4\% and 92.2\%, compared with 85.9\% for spaCy. However, spaCy achieves higher denied-span F1 (76.5\% versus 71.5\% and 73.3\%).

\begin{wraptable}{r}{0.45\textwidth}
\vspace{-2em}
\begin{minipage}{\linewidth}
\centering

\caption{Retrieval ablation with OpenAI QuickGELU CLIP. Entries are R-Neg@1/@5 (\%). V: number of prompt views; C: hybrid correction ($\lambda=0.3$, top-50 shortlist). Two views use rank fusion. Green values show gains over the first row.}
\label{tab:retrieval-decomposition}

\scriptsize

\begin{tabular}{@{}lccrr@{}}
\toprule
\textbf{Text} & \textbf{V} & \textbf{C}
& \textbf{COCO} & \textbf{MSR-VTT} \\
\midrule

Full caption & 1 & --
& 24.95 / 47.93 & 24.7 / 47.6 \\

Full caption & 2 & --
& 26.94 / 48.99 & 26.7 / 48.9 \\

\addlinespace[2pt]

Affirm. rem & 1 & --
& 29.41 / 53.55 & 27.0 / 52.8 \\

Affirm. rem & 2 & --
& 31.74 / 54.89 & 30.4 / 53.3 \\

Affirm. rem & 2 & Yes
& \textbf{33.27} / \textbf{56.40}
& \textbf{30.6} / \textbf{54.0} \\

\midrule

\multicolumn{3}{@{}l}{\textcolor{gaintext}{$\Delta$ improvement}}
& \textcolor{gaintext}{+8.32 / +8.47}
& \textcolor{gaintext}{+5.9 / +6.4} \\

\bottomrule
\end{tabular}
\end{minipage}
\par\vspace{0.8em}

\begin{minipage}{\linewidth}
\centering

\caption{Batch-one latency (ms) on an NVIDIA RTX A6000 with OpenAI CLIP ViT-B/32 QuickGELU in float32. P90 denotes the 90th percentile; the ratio compares median latencies.}
\label{tab:latency}

\scriptsize
\begin{tabular}{@{}lrrrrc@{}}
\toprule
& \multicolumn{2}{c}{\textbf{CLIP}}
& \multicolumn{2}{c}{\textbf{NAES}} & \\
\cmidrule(lr){2-3}
\cmidrule(lr){4-5}
\textbf{Dataset}
& Median & P90
& Median & P90
& \textbf{Ratio} \\
\midrule

\multicolumn{6}{@{}c}{\textit{Multiple-choice selection}} \\

COCO
& 9.10 & 9.39
& \cellcolor{latencybg}41.85
& \cellcolor{latencybg}42.62
& $4.6\times$ \\

VOC2007
& 8.47 & 8.53
& \cellcolor{latencybg}38.88
& \cellcolor{latencybg}39.75
& $4.6\times$ \\

\midrule

\multicolumn{6}{@{}c}{\textit{Retrieval}} \\

COCO
& 3.16 & 3.18
& \cellcolor{latencybg}5.27
& \cellcolor{latencybg}5.33
& $1.7\times$ \\

MSR-VTT
& 3.14 & 3.16
& \cellcolor{latencybg}5.23
& \cellcolor{latencybg}5.30
& $1.7\times$ \\

\bottomrule
\end{tabular}
\end{minipage}
\vspace{-5em}

\end{wraptable}

Despite these extraction differences, lexical, gold, and predicted decompositions yield identical recall on the annotated COCO retrieval subset; on MSR-VTT, recall varies modestly across decomposition variants, with no clear advantage for any approach (Appendix~\ref{app:decomposer}). Thus, the downstream results support the decomposition-based scoring strategy but do not establish that learned decomposition is necessary for these benchmarks. Annotations use single annotators on disjoint sets, so inter-annotator agreement is unavailable.

\subsection{Qualitative results}
\label{sec:qualitative}

\begin{figure}[!h]
\centering
\includegraphics[width=0.95\linewidth]{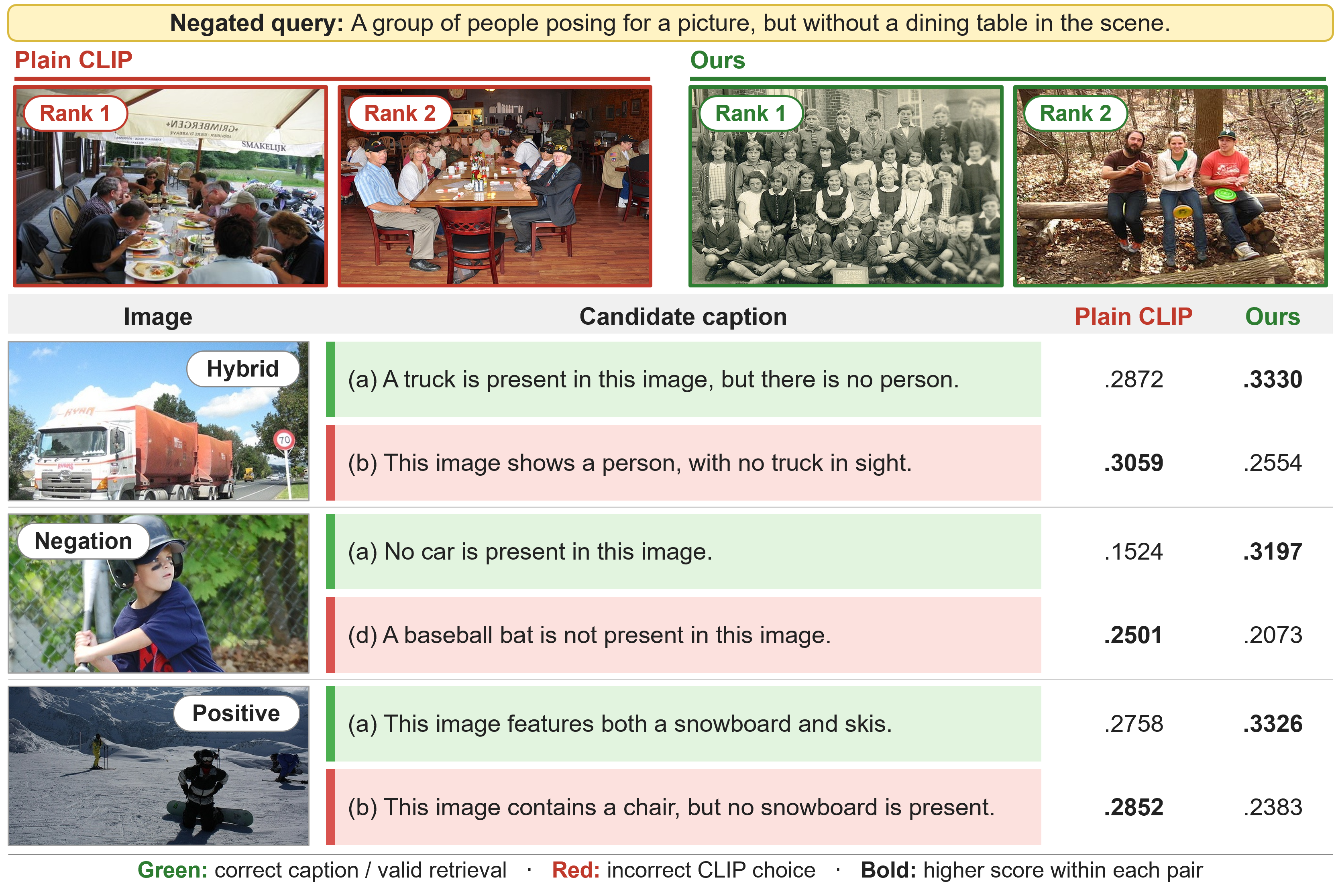}
\caption{Qualitative retrieval and MCQ examples. Top: for a query negating ``dining table,'' plain CLIP's top-2 retrievals both contain one, while ours excludes it at both ranks. Illustrative successes, not representative results. Bottom: three MCQ corrections spanning hybrid, negation-only, and positive answers; green marks the correct caption, red marks plain CLIP's top choice.}
\label{fig:qualitative}
\end{figure}

\begin{figure}[!h]
\centering
\includegraphics[width=\linewidth,height=\textheight,keepaspectratio]{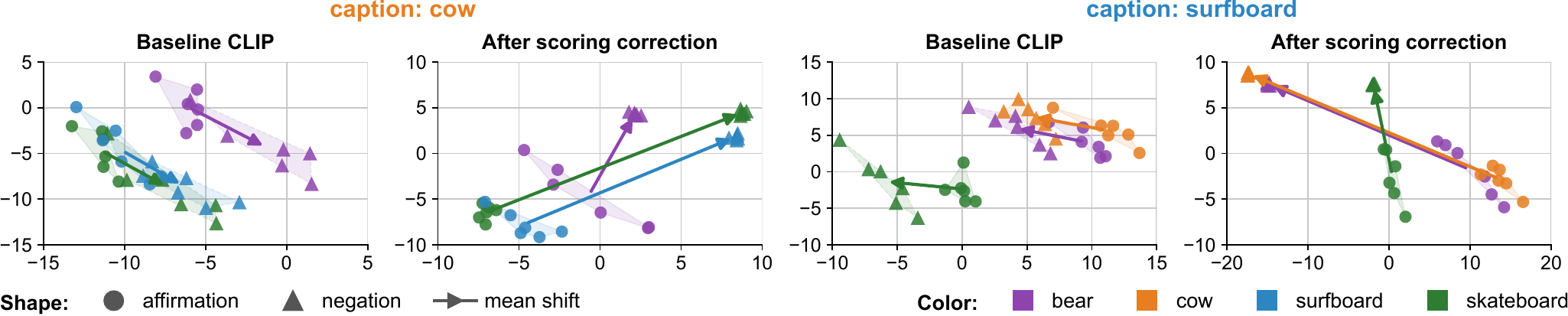}
\caption{Shared PCA of affirmative and negated scoring profiles under baseline CLIP and candidate-conditioned NAES with crop grounding (OpenAI QuickGELU). Profiles encode negative ordinal ranks averaged over two scoring views across 256 COCO images, using six parser-validated templates per object pair. A single PCA spans all four objects (bear, cow, surfboard, skateboard; PC1: 27.6\%, PC2: 11.8\%), revealing greater polarity separation after correction.}
\label{fig:scoring-profile-pca}
\end{figure}

Figure~\ref{fig:qualitative} illustrates ranking changes in retrieval and three corrected MCQ cases. Figure~\ref{fig:scoring-profile-pca} complements these with a shared PCA visualization, showing greater separation between affirmative and negated scoring after correction. Appendix~\ref{app:qualify} examines failures.

\subsection{Ablations}
\label{sec:ablation}

\paragraph{Retrieval components:} Affirmative decomposition and prompt fusion provide complementary gains (Table~\ref{tab:retrieval-decomposition}). With one template, replacing the full caption with its affirmative remainder increases COCO R-Neg@1 from 24.95\% to 29.41\%. Two-view fusion raises it further to 31.74\%. Fusion also improves full-caption retrieval, suggesting a general benefit from combining prompt-dependent rankings. The hybrid correction adds a smaller improvement,
reaching 33.27\%.

The same pattern holds on MSR-VTT: decomposition and fusion together raise R-Neg@1 from 24.7\% to 30.4\%, while correction adds only 0.2 points. Paired intervals support decomposition-plus-fusion gains on both datasets; for OpenAI COCO R-Neg@1, the gain is $+6.79$ points with a 95\% interval of $[+6.33,+7.25]$. The incremental correction's MSR-VTT intervals include
zero (Appendix~\ref{app:correction}). Most retrieval improvement, therefore, comes from the affirmative reformulation and rank aggregation.

\paragraph{MCQ components:} Table~\ref{tab:crop-ablation} distinguishes better visual localization from polarity-aware scoring. Lexical crop grounding alone changes overall accuracy by only $+1.5$ points on COCO and $-0.4$ on VOC: locating mentioned objects does not determine whether the caption affirms or denies them. Introducing the affirmative remainder adds 11.9 and 14.2 points, respectively, but leaves negation-only accuracy at 0.1\%. Independent reflection addresses this missing case, adding 16.3 and 28.6 points overall.

\begin{table}[!h]
\centering
\caption{Component ablation for OpenAI CLIP with the from-scratch decomposer and fixed $\lambda=0.3$ where applicable. Accuracy (\%) is averaged over 24 option permutations. $\Delta$ is the overall accuracy change from the preceding row (percentage points), with paired image-cluster 95\% confidence intervals. Bold indicates the best accuracy in each column. A/B use full captions with two-view rank fusion. D2 replaces D's external reference with the candidate-pool mean; E uses the full candidate-conditioned scoring rule.}
\label{tab:crop-ablation}

\setlength{\tabcolsep}{3pt}
\renewcommand{\arraystretch}{1.2}

\resizebox{\linewidth}{!}{%
\begin{tabular}{@{}cl*{5}{r}c*{5}{r}c@{}}
\toprule
& & \multicolumn{6}{c}{\textbf{COCO-MCQ}}
    & \multicolumn{6}{c}{\textbf{VOC2007-MCQ}} \\
\cmidrule(lr){3-8}
\cmidrule(lr){9-14}

\textbf{ID} & \textbf{Configuration}
& Pos. & Hyb. & Neg. & \textbf{Avg}
& $\boldsymbol{\Delta}$ & \textbf{95\% CI}
& Pos. & Hyb. & Neg. & \textbf{Avg}
& $\boldsymbol{\Delta}$ & \textbf{95\% CI} \\
\midrule

\rowcolor{absection}
\multicolumn{14}{@{}c@{}}{\textbf{Visual representation}} \\

A & Global image, full caption
& 68.0 & 42.6 & 6.8
& 40.0
& -- & --
& 81.5 & 59.5 & 2.8
& 38.6
& -- & -- \\

B & + Lexical crop grounding
& 73.4 & 45.2 & 3.0
& 41.6
& \textcolor{gaintext}{+1.5} & $[+0.5,+2.5]$
& 84.5 & 58.6 & 1.9
& 38.2
& $-0.4$ & $[-1.3,+0.5]$ \\

\addlinespace[3pt]

\rowcolor{absection}
\multicolumn{14}{@{}c@{}}{\textbf{Caption decomposition and correction}} \\

C0 & + Affirmative remainder
& \textbf{82.2} & 73.8 & 0.1
& 53.4
& \textcolor{gaintext}{+11.9} & $[+10.9,+12.8]$
& 90.5 & \textbf{90.6} & 0.1
& 52.4
& \textcolor{gaintext}{+14.2} & $[+13.2,+15.2]$ \\

C & + Hybrid correction
& \textbf{82.2} & \textbf{75.7} & 0.1
& 54.0
& \textcolor{gaintext}{+0.6} & $[+0.3,+0.9]$
& \textbf{90.6} & 90.4 & 0.1
& 52.3
& $-0.1$ & $[-0.3,+0.2]$ \\

\addlinespace[3pt]

\rowcolor{absection}
\multicolumn{14}{@{}c@{}}{\textbf{Negation-only scoring}} \\

D & + Independent reflection
& 76.8 & 65.8 & 68.1
& 70.3
& \textcolor{gaintext}{+16.3} & $[+15.1,+17.5]$
& 88.7 & 87.9 & 71.3
& 81.0
& \textcolor{gaintext}{+28.7} & $[+27.3,+30.0]$ \\

D2 & Pool-centered reflection
& 74.5 & 73.7 & 72.1
& 73.5
& \textcolor{gaintext}{+3.2} & $[+2.4,+4.0]$
& 87.4 & 89.9 & 86.7
& 88.2
& \textcolor{gaintext}{+7.2} & $[+6.3,+8.1]$ \\

\cmidrule(lr){2-14}

E & \textbf{Candidate-conditioned (full)}
& 76.5 & 75.5 & \textbf{84.4}
& \textbf{78.7}
& \textcolor{gaintext}{+5.2} & $[+4.6,+5.8]$
& 86.4 & 90.4 & \textbf{94.2}
& \textbf{91.5}
& \textcolor{gaintext}{+3.2} & $[+2.7,+3.8]$ \\

\bottomrule
\end{tabular}}
\end{table}

Candidate conditioning provides further improvement. Replacing the external reference with the pool mean adds 3.2 points on COCO and 7.3 on VOC; the full mapping adds another 5.1 and 3.1 points, with confidence intervals entirely above zero. Thus, a pool-dependent center explains only part of the gain: claim aggregation and score alignment also contribute.

\paragraph{Encoder transfer:} Improvements extend beyond the primary ViT-B/32 encoders (Appendix~\ref{app:backbone-transfer}, Table~\ref{tab:backbone-transfer}). For SigLIP~2 ViT-L/16, COCO-MCQ accuracy increases from 25.1\% to 84.8\%, while COCO R-Neg@5 rises from 70.8\% to 77.5\%. The consistent improvement across SigLIP, SigLIP~2, \& larger CLIP encoders suggests the benefit is not specific to one pretrained representation. Its magnitude, nevertheless, depends on the encoder
and task.

\section{Limitations}
\label{sec:limitations}

\paragraph{Inference overhead:} NAES adds caption decomposition, crop encoding, and evidence queries. On our NVIDIA RTX A6000 setup, batch-one inference is approximately $4.6\times$ slower than CLIP for MCQ and $1.7\times$ slower for retrieval with a cached gallery (Table~\ref{tab:latency}). These measurements exclude initialization, media decoding, gallery construction, and detector-backed VALSE inference; the overhead may differ under other hardware and batching configurations.

\paragraph{Candidate-set dependence:} Candidate-conditioned scoring improves LAION-2B COCO-MCQ accuracy from 74.6\% to 82.4\%, but it uses the competing options to center and normalize evidence. Accuracy falls by approximately 9--16 percentage points with six or eight options, and by up to 18.7 points when a confirmed-present object is added to a hybrid distractor (Appendix~\ref{app:pool}). These tests change both the scoring context and the difficulty of the alternatives, so they do not isolate normalization effects. Establishing ranking stability requires controlled tests that hold the image and focal captions fixed while varying unrelated distractors. Pool-independent scoring remains available when access to competing descriptions is undesirable.

\paragraph{Scope of presence evidence:} NAES uses frozen encoder similarities as evidence for mentioned objects. Weak responses can reflect small size, occlusion, or recognition failure rather than absence; we have not separately quantified these factors. Moreover, object presence alone does not establish attributes, events, or relations, as examined on NegRefCOCOg (Appendix~\ref{app:negref}), and NegSplit's own parsing may rely on surface cues rather than deeper structure (Appendix~\ref{app:cue}). Our results show better description ranking on the evaluated benchmarks. Independent verification of arbitrary negated claims remains untested.

\section{Conclusion}
\label{sec:conclusion}

We introduced Negation-Aware Evidence Scoring (NAES), which separates affirmative matching from denied-object evidence while keeping vision–language encoders frozen. Across multiple-choice selection and image and video retrieval, the results show that reformulating queries and combining evidence according to polarity can substantially improve negation-aware ranking. Ablations identify complementary contributions: affirmative decomposition and prompt fusion explain most retrieval gains, while presence-score reflection and candidate-conditioned score alignment address negation-only selection. The central finding is that useful affirmative evidence can support better handling of negation without retraining the underlying encoders. Extending this benefit to reliable verification requires addressing the distinction between weak presence evidence and absence, dependence on competing descriptions, and negated attributes, events, and relations.

\bibliography{main}
\bibliographystyle{tmlr}

\appendix

\section{Scoring Details}
\label{app:scoring}

\paragraph{Crop-grounded image vectors for MCQ:} Let $\mathcal R(I)$ contain the whole image, its four non-overlapping $2\times2$ cells, and its nine non-overlapping $3\times3$ cells, and let $v_{I,r}$ be the $\ell_2$-normalized frozen CLIP feature for crop $r$. For a category $o$, let $p_m(o)$ be the four fixed prompts (``a photo of $o$,'' ``a picture of $o$,'' ``an image showing $o$,'' and ``$o$ in a photo''). Its concept vector is $q_o=\operatorname{norm}(\frac{1}{4}\sum_{m=1}^4 f_T(p_m(o)))$, where $f_T$ returns an $\ell_2$-normalized CLIP text feature and $\operatorname{norm}(x)=x/\lVert x\rVert_2$. For the fixed 80-category vocabulary $V$, object-specific attention and crop aggregation are
\begin{equation}
 \alpha_{I,r,o}=\frac{\exp(\kappa\langle v_{I,r},q_o\rangle)}{\sum_{u\in V}\exp(\kappa\langle v_{I,r},q_u\rangle)},\quad
 g_{I,o}=\operatorname{norm}\!\Big(\sum_{r\in\mathcal R(I)}\alpha_{I,r,o}v_{I,r}\Big),\quad
 g_{I,P_k}=\operatorname{norm}\!\Big(\sum_{o\in P_k}g_{I,o}\Big),
 \label{eq:grounded-image}
\end{equation}
where $P_k$ is the set of distinct affirmed categories recognized in option $k$ by the fixed matcher, and $\kappa=\exp(\texttt{logit\_scale})$ is CLIP's frozen logit scale. If $P_k$ is empty, the implementation uses the normalized sum of the whole-image and horizontal-flip image features. Retrieval keeps the global gallery image or video feature. MCQ views are $A_{v,k}=\langle g_{I,P_k},f_T(\rho_v(x_k))\rangle$.

\paragraph{Hybrid correction details:} The correction uses $n(t_k)$, NegSplit's open-vocabulary denied phrase, which is free text and not restricted to the 80-category matcher used by localized reflection. $\mu_A,\sigma_A$ are per-fold moments of native hybrid scores. For MCQ, we estimated them from the training folds; for retrieval, we use the per-view mean and sample standard deviation of native top-50 scores over resolved hybrid queries in the other four folds (COCO folds are image-disjoint, and MSR-VTT folds are query-disjoint). For retrieval, the correction is applied within each hybrid query's native top-50 shortlist before full-gallery rank fusion. Unresolved hybrid candidates keep $A_{v,k}$ unchanged. A hybrid and a positive candidate with identical affirmative content can therefore tie when the hybrid is unresolved or its correction is zero.

\paragraph{Independent reflection for caption-foil pairs:} VALSE uses a global-evidence backend. One caption in the pair is the unmodified positive counterpart, so its native CLIP score $s_0$ is reused as the reference. For a single denied concept,
\begin{equation}
 A^{\mathrm{VALSE}}_{\mathrm{positive}}=s_0,\qquad
 A^{\mathrm{VALSE}}_{\mathrm{negative}}=s_0+2(c_{\mathrm{global}}-e_o^{\mathrm{global}}).
 \label{eq:ecpr-pair}
\end{equation}
This corrects only the negation-only caption. For detector evidence, $e_o$ is the maximum confidence for the denied phrase with its own boundary ($c=0.06$ for OWL-ViT, from a separate 900-score COCO calibration), and no threshold is fitted on VALSE. VALSE uses the denied phrase truncated before its first preposition or participle, since detector queries need a bare concept name.

\paragraph{Candidate-conditioned fallbacks:} If there is one candidate, fewer than two distinct concepts, incomplete evidence, or $\sigma_C\le\epsilon$, the route falls back to pool-independent reflection when the denied concepts resolve, and otherwise retains the native score. $P_k$ and $N_k$ come from the same 80-category matcher, so an unresolved concept outside the vocabulary is absent from both sets rather than falling back to other evidence. Averaging the signed terms gives every resolved claim equal weight regardless of caption length.

\paragraph{Tie handling and permutations:} Within each MCQ view, exact ties are broken by the original candidate order, and ties in mean rank go to the earliest candidate. We evaluate the rule over all 24 permutations of the four options and report the uniform mean. We keep original-order accuracy as a supplemental diagnostic. For retrieval, recall at $K$ counts as a hit when the paired target appears among the first $K$ gallery items; this applies to both R@1 and R@5.

\FloatBarrier
\section{Calibration of External References}
\label{app:calibration}

The global boundary $c_{\mathrm{global}}$ is the zero-logit threshold of a logistic presence fit on CLIP's whole-image cosine scores, using 300,000 CC3M images and OWL-ViT-derived presence and absence pseudo-labels~\citep{sharma2018cc3m,minderer2022owlvit}. The values are $c_{\mathrm{global}}=0.214727$ for OpenAI QuickGELU and $0.146967$ for LAION-2B. The OpenAI boundary uses the same QuickGELU encoder as the benchmark results, uses no COCO or VOC answer labels, and does not refit $\lambda$. Refitting on 20,000, 150,000, and 300,000 images gives $c=0.234443$, $0.216871$, and $0.214727$, with held-out AUROCs of $0.9269$, $0.9157$, and $0.8983$. The 20,000-image sample is imbalanced, and each size uses its own held-out pairs (Figure~\ref{fig:cc3mroc}). The threshold stabilizes with sample size: $c$ shifts by $0.0176$ from 20,000 to 150,000 images but only $0.0021$ from 150,000 to 300,000, while discrimination remains strong throughout; we use the 300,000-image fit for this more stable estimate. The same stabilization pattern holds for LAION-2B, which follows the same procedure.

\begin{figure}[!h]
\centering
\includegraphics[width=\linewidth]{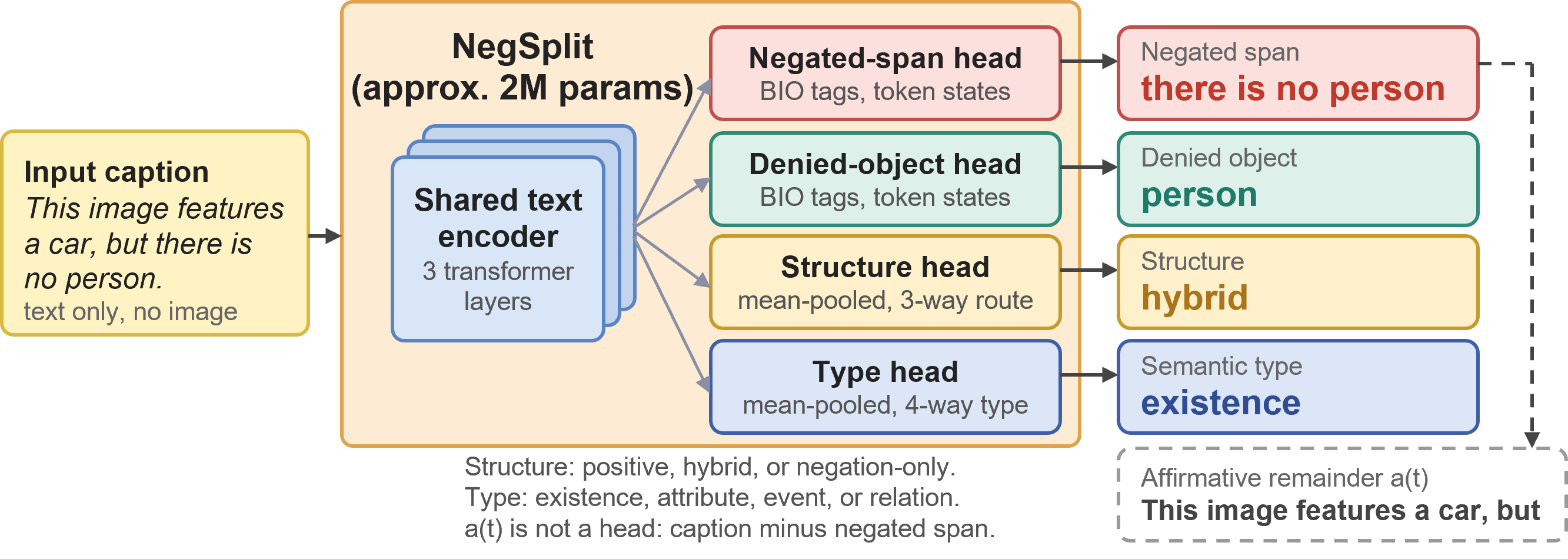}
\caption{NegSplit's architecture (from-scratch transformer). A shared text encoder feeds two BIO heads (negated span, denied content) and, via its mean-pooled state, the structure and semantic-type heads. The affirmative remainder $a(t)$ (dashed) is the caption minus the decoded negated span, not predicted directly. Spans use longest-contiguous-run decoding; the example is illustrative.}
\label{fig:negsplit}
\end{figure}

Localized MCQ reflection uses a separate calibration. Crop-centered category evidence is fit on the CC3M calibration split and affinely mapped to the score scale of the crop-grounded object embeddings, preserving the fitted logit. The references are $c_b=0.267216$ for OpenAI and $0.246679$ for LAION-2B. Its CC3M calibration and evaluation images are disjoint. VALSE's detector rows use the separate OWL-ViT boundary of $0.06$.

\section{Setup Details}
\label{app:setup}

\paragraph{NegSplit:} NegSplit trains on 181,734 synthetic text examples generated from CC3M captions and cue families for existence, attribute, event, and relation negation, balancing the positive, hybrid, and negation-only routes at 60,578 examples each. The encoder options (Figure~\ref{fig:negsplit}) are a from-scratch transformer (1,943,823 parameters: hidden size 128, three layers, four heads, feed-forward width 256, learning rate $10^{-3}$) and a pruned BERT-tiny encoder~\citep{turc2019berttiny} (2,000,527 parameters, learning rate $10^{-4}$). Two token-level BIO heads predict the negated and denied-object spans, with the final span chosen as the longest unbroken run of tagged tokens. A mean-pooled sequence head predicts the structure, and a fourth head predicts the semantic type. All heads train jointly with a summed loss. We evaluate the same fixed checkpoints across all experiments, and within each task both checkpoints share the same scoring rules, gallery, evidence backend, external boundary, and evaluation population.

\begin{table}[!htbp]
\centering

\parbox[t]{0.48\linewidth}{%
    \vspace{0pt}
    \centering
    \caption{Candidate-conditioned NAES accuracy (\%), averaged over 24 option orders. We select $\lambda=0.3$ on LAION-2B COCO retrieval and transfer it unchanged. Bold marks the best accuracy.}
    \label{tab:coco-voc-transfer}

    \scriptsize
    \setlength{\tabcolsep}{3pt}
    \renewcommand{\arraystretch}{1.10}
    \begin{adjustbox}{max width=\linewidth}
    \begin{tabular}{@{}lrr@{}}
    \toprule
    \textbf{Decomposer} & \textbf{COCO-MCQ} & \textbf{VOC2007-MCQ} \\
    \midrule
    \rowcolor{backbonebg}
    \multicolumn{3}{@{}c@{}}{\textbf{OpenAI CLIP}} \\
    From scratch & \textbf{78.69} & \textbf{91.48} \\
    BERT-tiny    & 78.40 & 90.86 \\
    \midrule
    \rowcolor{backbonebg}
    \multicolumn{3}{@{}c@{}}{\textbf{LAION-2B CLIP}} \\
    From scratch & \textbf{82.42} & \textbf{93.02} \\
    BERT-tiny    & 82.14 & 92.41 \\
    \bottomrule
    \end{tabular}
    \end{adjustbox}\par
}%
\hfill
\parbox[t]{0.48\linewidth}{%
    \vspace{0pt}
    \centering
    \caption{Candidate-conditioned accuracy (\%) by recognized affirmed-object count, with $\lambda=0.3$ and 24-order averaging. Shading is descriptive.}
    \label{tab:affirmed-count}

    \scriptsize
    \setlength{\tabcolsep}{3pt}
    \renewcommand{\arraystretch}{1.10}
    \begin{adjustbox}{max width=\linewidth}
    \begin{tabular}{@{}crrr@{}}
    \toprule
    \textbf{Objects} & $\boldsymbol{n}$
    & \textbf{LAION-2B} & \textbf{OpenAI} \\
    \midrule
    \rowcolor{backbonebg}
    \multicolumn{4}{@{}c@{}}{\textbf{COCO-MCQ}} \\
    0  & 1,865 & 87.75 & 84.61 \\
    1  & 1,990 & 79.17 & 75.47 \\
    2  & 2,056 & 80.78 & 76.51 \\
    \rowcolor{sparsebg}
    3+ &     3 & 50.00 & 33.33 \\
    \midrule
    \rowcolor{backbonebg}
    \multicolumn{4}{@{}c@{}}{\textbf{VOC2007-MCQ}} \\
    0  & 2,124 & 95.50 & 94.40 \\
    1  & 2,196 & 91.20 & 90.52 \\
    2  &   710 & 91.29 & 85.83 \\
    \rowcolor{sparsebg}
    3+ &     1 & 33.33 &  0.00 \\
    \bottomrule
    \end{tabular}
    \end{adjustbox}\par
}
\end{table}

\section{Additional Supporting Results}
\label{app:support}

We provide protocol details for evidence and retrieval, a visualization of scoring structure, and extended results for MCQ, backbone transfer, retrieval, visual-dependence controls, VALSE-Existence, decomposer evaluation, and inference cost.

\paragraph{Evidence and retrieval protocol:} Detector evidence comes from \texttt{google/owlvit-base-patch32}~\citep{minderer2022owlvit} and is used only for VALSE. MCQ uses candidate-specific crop-grounded vectors and crop-centered reflection evidence, whereas the hybrid correction uses a global image vector with a free-text denied query. MSR-VTT uses four uniformly sampled frames per video, mean-pooled before normalization. The two retrieval text views are the unwrapped description and \texttt{a photo of \{description\}}.

\paragraph{Visualization of scoring structure:} Figure~\ref{fig:scoring-profile-pca} shows shared-PCA profiles of affirmative and negated captions before and after scoring correction. Corrected negation profiles form tighter groups with clear affirmation-negation separation, and their location depends on the companion context.

\subsection{Additional MCQ results}
\label{app:mcq-extra}

Table~\ref{tab:coco-voc-transfer} gives the fixed-weight transfer across decomposers and backbones; Table~\ref{tab:ablation} provides NAES's pool-independent and candidate-conditioned components with the BERT-tiny decomposer checkpoint for comparison with Table~\ref{tab:crop-ablation}'s scratch-decomposer rows D and E, and Table~\ref{tab:affirmed-count} stratifies accuracy by the number of recognized affirmed objects. Strata use predicted NegSplit spans, and correct answers are used only for grouping; the zero stratum includes negation-only options and unrecognized affirmed objects.

\begin{table}[!h]
\centering
\caption{Reflection ablation with BERT-tiny NegSplit, OpenAI CLIP, and fixed $\lambda=0.3$. Accuracy (\%) averaged over 24 orders. D/E correspond to the scratch-decomposer configurations in Table~\ref{tab:crop-ablation}. $\Delta$ reports E minus D in percentage points; green text marks increases. Bold identifies the higher accuracy per metric.}
\label{tab:ablation}

\begin{tabular}{@{}l*{8}{r}@{}}
\toprule
& \multicolumn{4}{c}{\textbf{COCO-MCQ}}
& \multicolumn{4}{c}{\textbf{VOC2007-MCQ}} \\
\cmidrule(lr){2-5}
\cmidrule(lr){6-9}
\textbf{Scoring}
& Pos. & Hyb. & Neg. & \textbf{Overall}
& Pos. & Hyb. & Neg. & \textbf{Overall} \\
\midrule

D\quad Independent
& \textbf{76.3} & 66.5 & 67.4
& \cellcolor{aboverall}70.1
& \textbf{88.3} & 88.2 & 69.1
& \cellcolor{aboverall}80.2 \\

E\quad Candidate-conditioned
& 76.0 & \textbf{75.8} & \textbf{84.0}
& \cellcolor{aboverall}\textbf{78.5}
& 85.9 & \textbf{90.5} & \textbf{93.2}
& \cellcolor{aboverall}\textbf{91.0} \\

\midrule

$\Delta$ (E $-$ D)
& $-0.3$
& \textcolor{gaintext}{+9.3}
& \textcolor{gaintext}{+16.6}
& \textcolor{gaintext}{+8.4}
& $-2.4$
& \textcolor{gaintext}{+2.3}
& \textcolor{gaintext}{+24.1}
& \textcolor{gaintext}{+10.8} \\

\bottomrule
\end{tabular}
\end{table}

\subsection{Backbone transfer}
\label{app:backbone-transfer}
Table~\ref{tab:backbone-transfer} gives the full per-encoder results summarized in the ``Encoder transfer'' paragraph of Section~\ref{sec:results}.

\newcommand{\transfergain}[1]{\textcolor{gaintext}{+#1}}
\newcommand{\transferscore}[1]{\cellcolor{transferbg}#1}

\begin{table}[!h]
\centering
\caption{Transfer across frozen vision-language encoders. Entries are percentages. Lavender cells report NAES scores; green values indicate absolute gains over the original encoder in percentage points. MCQ accuracy is averaged over 24 option orders. Original-caption R@5 is unchanged by correction. Bold marks the best NAES scores and original-caption recalls.}
\label{tab:backbone-transfer}

\setlength{\tabcolsep}{3pt}
\renewcommand{\arraystretch}{1.15}

\resizebox{\linewidth}{!}{%
\begin{tabular}{@{}l*{14}{r}@{}}
\toprule
\multirow{3}{*}{\textbf{Encoder}}
& \multicolumn{3}{c}{\textbf{COCO-MCQ}}
& \multicolumn{3}{c}{\textbf{VOC2007-MCQ}}
& \multicolumn{4}{c}{\textbf{COCO retrieval}}
& \multicolumn{4}{c}{\textbf{MSR-VTT retrieval}} \\
\cmidrule(lr){2-4}
\cmidrule(lr){5-7}
\cmidrule(lr){8-11}
\cmidrule(lr){12-15}

& \multirow{2}{*}{Orig.}
& \multirow{2}{*}{NAES}
& \multirow{2}{*}{$\Delta$}
& \multirow{2}{*}{Orig.}
& \multirow{2}{*}{NAES}
& \multirow{2}{*}{$\Delta$}
& \multirow{2}{*}{R@5}
& \multicolumn{3}{c}{R-Neg@5}
& \multirow{2}{*}{R@5}
& \multicolumn{3}{c}{R-Neg@5} \\
\cmidrule(lr){9-11}
\cmidrule(lr){13-15}
& & & & & & &
& Orig. & NAES & $\Delta$
& & Orig. & NAES & $\Delta$ \\
\midrule

\rowcolor{backbonebg}
\multicolumn{15}{@{}c@{}}{\textbf{SigLIP encoders - WebLI}} \\

SigLIP ViT-B/16
& 28.9 & \transferscore{81.2} & \transfergain{52.3}
& 30.8 & \transferscore{89.7} & \transfergain{58.9}
& 72.1 & 64.4 & \transferscore{72.2} & \transfergain{7.8}
& 53.6 & 48.2 & \transferscore{54.0} & \transfergain{5.8} \\

SigLIP 2 ViT-B/32
& 27.2 & \transferscore{84.3} & \transfergain{57.1}
& 27.1 & \transferscore{93.9} & \transfergain{66.8}
& 72.8 & 64.0 & \transferscore{72.3} & \transfergain{8.3}
& 56.3 & 49.6 & \transferscore{55.5} & \transfergain{5.9} \\

SigLIP 2 ViT-L/16
& 25.1 & \transferscore{\textbf{84.8}} & \transfergain{59.7}
& 24.5 & \transferscore{\textbf{94.2}} & \transfergain{69.7}
& \textbf{76.9} & 70.8
& \transferscore{\textbf{77.5}} & \transfergain{6.7}
& \textbf{63.4} & 54.4
& \transferscore{\textbf{62.6}} & \transfergain{8.2} \\

\midrule
\rowcolor{backbonebg}
\multicolumn{15}{@{}c@{}}{\textbf{CLIP encoders}} \\

ViT-H/14 (LAION-2B)
& 30.3 & \transferscore{82.8} & \transfergain{52.5}
& 32.1 & \transferscore{91.1} & \transfergain{59.0}
& 72.8 & 66.3 & \transferscore{72.4} & \transfergain{6.1}
& 58.7 & 54.2 & \transferscore{59.7} & \transfergain{5.5} \\

ViT-L/14 (OpenAI)
& 38.0 & \transferscore{81.2} & \transfergain{43.2}
& 39.2 & \transferscore{91.6} & \transfergain{52.4}
& 60.0 & 52.3 & \transferscore{61.3} & \transfergain{9.0}
& 57.1 & 51.7 & \transferscore{58.3} & \transfergain{6.6} \\

ViT-L/14 (LAION-2B)
& 25.5 & \transferscore{81.8} & \transfergain{56.3}
& 25.3 & \transferscore{91.8} & \transfergain{66.5}
& 70.4 & 63.4 & \transferscore{70.5} & \transfergain{7.1}
& 57.5 & 53.3 & \transferscore{59.0} & \transfergain{5.7} \\

\bottomrule
\end{tabular}}
\end{table}

\FloatBarrier
\subsection{Additional retrieval results}
\label{app:retrieval}

Table~\ref{tab:retrieval-uncertainty} reports the paired bootstrap for the pipeline before the hybrid correction.

\begin{wraptable}{r}{0.52\linewidth}
\centering
\setlength{\tabcolsep}{3pt}
\renewcommand{\arraystretch}{1.10}
\caption{Retrieval gains over matched CLIP before hybrid correction, using the scratch decomposer and two-view fusion. Before/After are full-caption and decomposition-plus-fusion recall (\%); $\Delta$ is in percentage points. Paired 95\% intervals use 10,000 bootstrap resamples of source images (COCO) or videos (MSR-VTT) on the full benchmarks.}
\label{tab:retrieval-uncertainty}
{\scriptsize
\begin{tabular}{@{}llrrrc@{}}
\toprule
\textbf{Dataset} & \textbf{Metric}
& \textbf{Before} & \textbf{After} & $\boldsymbol{\Delta}$ & \textbf{95\% CI} \\
\midrule
COCO & R-Neg@1
& 24.95 & 31.74
& \textcolor{gaintext}{+6.79}
& $[+6.33,+7.25]$ \\
& R-Neg@5
& 47.93 & 54.89
& \textcolor{gaintext}{+6.95}
& $[+6.48,+7.42]$ \\
\midrule
MSR-VTT & R-Neg@1
& 24.7 & 30.4
& \textcolor{gaintext}{+5.70}
& $[+3.70,+7.80]$ \\
& R-Neg@5
& 47.6 & 53.3
& \textcolor{gaintext}{+5.70}
& $[+3.60,+7.90]$ \\
\bottomrule
\end{tabular}
\par}
\end{wraptable}

\FloatBarrier
\subsection{Visual-dependence controls}
\label{app:controls}

Table~\ref{tab:visual-controls} tests whether each scoring signal depends on the evaluated image, decomposed by answer type. Template ensembling explains little of the gap between native CLIP and the full method: Templated CLIP stays within 1--2 points of Plain CLIP under both the correct and the wrong image, on both datasets. The two scoring signals are cleanly separable, with no measurable cross-talk: corrupting grounding alone collapses positive and hybrid accuracy while leaving negation-only accuracy unchanged; corrupting concept evidence alone collapses negation-only accuracy while leaving positive and hybrid accuracy unchanged (within approximately 1.7 points on every comparison). This is direct evidence that NAES's negation-only correction uses real, image-specific evidence rather than an image-agnostic prior, isolated to exactly the answer type it serves.

\begin{table}[!h]
\centering
\caption{Visual-dependence controls for OpenAI CLIP (from-scratch decomposer, $\lambda=0.3$). Real/Donor/Zero mark the image source for each signal: Matching gives affirmative image--text scores, Evidence gives category-presence scores. Templated CLIP applies two-view fusion to full captions. Donor rows average five seeds; no-image rows are deterministic. Lavender marks intact NAES accuracy; yellow marks negation-only accuracy under corrupted matching with intact evidence.}
\label{tab:visual-controls}

\resizebox{\linewidth}{!}{%
\begin{tabular}{@{}lcc*{8}{r}@{}}
\toprule
& \multicolumn{2}{c}{\textbf{Image source}}
& \multicolumn{4}{c}{\textbf{COCO-MCQ}}
& \multicolumn{4}{c}{\textbf{VOC2007-MCQ}} \\
\cmidrule(lr){2-3}
\cmidrule(lr){4-7}
\cmidrule(lr){8-11}

\textbf{Scorer}
& Matching & Evidence
& Pos. & Hyb. & Neg. & \textbf{Overall}
& Pos. & Hyb. & Neg. & \textbf{Overall} \\
\midrule

\rowcolor{backbonebg}
\multicolumn{11}{@{}c@{}}{\textbf{CLIP controls}} \\

CLIP & Real & --
& 69.1 & 39.3 & 6.8 & 39.3
& 82.6 & 59.0 & 3.3 & 38.7 \\

Templated CLIP & Real & --
& 68.0 & 42.7 & 6.8 & 40.0
& 81.5 & 59.5 & 2.8 & 38.6 \\

\addlinespace[2pt]

CLIP & Donor & --
& 19.2 & 16.1 & 17.6 & 17.7
& 21.6 & 15.6 & 16.4 & 16.7 \\

Templated CLIP & Donor & --
& 17.5 & 18.5 & 17.9 & 18.0
& 18.7 & 16.4 & 18.1 & 17.4 \\

\midrule

\rowcolor{backbonebg}
\multicolumn{11}{@{}c@{}}{\textbf{NAES: intact and corrupted signals}} \\

NAES & Real & Real
& 76.6 & 75.5 & 84.5
& \cellcolor{controlbg}78.7
& 86.4 & 90.4 & 94.2
& \cellcolor{controlbg}91.5 \\

\cmidrule(lr){2-11}

NAES & Donor & Real
& 21.0 & 42.3
& \cellcolor{retainedbg}84.3
& 48.3
& 25.4 & 43.4
& \cellcolor{retainedbg}93.7
& 62.1 \\

NAES & Real & Donor
& 75.1 & 75.9 & 48.6 & 67.0
& 84.7 & 90.6 & 46.4 & 71.2 \\

NAES & Donor & Donor
& 20.4 & 42.2 & 48.7 & 36.7
& 24.6 & 43.2 & 46.2 & 41.9 \\

\cmidrule(lr){2-11}

NAES & Zero & Zero
& 0.0 & 0.5 & 50.0 & 16.0
& 0.0 & 0.7 & 49.8 & 21.3 \\

\bottomrule
\end{tabular}}
\end{table}

\begin{table}[!h]
\centering
\caption{Accuracy (\%) on VALSE-Existence and NegRefCOCOg. Measured results use the configurations in Section~\ref{sec:setup}. The clause parser is a dataset-specific reference; yellow marks external detector evidence. Bold is the best per benchmark.}
\label{tab:valse}

\begin{tabular}{@{}llrr@{}}
\toprule
\textbf{Method / scoring}
& \textbf{Evidence / adapter}
& \textbf{VALSE}
& \textbf{NegRefCOCOg} \\
\midrule

\rowcolor{backbonebg}
\multicolumn{4}{@{}c@{}}{\textbf{Baselines and references}} \\

CLIP (matched)
& Native cosine
& 69.3 & 58.0 \\

PeakPatch~\citep{lu2026peakpatch}
& Published result
& 80.4 & 64.3 \\

Aggarwal et al.~\citep{aggarwal2026seeingwhat}
& Published result
& 79.5 & -- \\

Clause-parser contrast
& Hand-built reference
& -- & \textbf{67.3} \\

\midrule

\rowcolor{backbonebg}
\multicolumn{4}{@{}c@{}}{\textbf{NegSplit - from scratch}} \\

Independent reflection
& CLIP
& 77.6 & -- \\

Counterpart anchoring
& CLIP
& 75.2 & -- \\

Counterpart anchoring
& \cellcolor{detectorbg}OWL-ViT detector
& 82.2 & -- \\

One-template contrast
& Affirmative minus counterfactual
& -- & 60.7 \\

\midrule

\rowcolor{backbonebg}
\multicolumn{4}{@{}c@{}}{\textbf{NegSplit - BERT-tiny}} \\

Independent reflection
& CLIP
& 77.0 & -- \\

Counterpart anchoring
& CLIP
& 75.4 & -- \\

Counterpart anchoring
& \cellcolor{detectorbg}OWL-ViT detector
& \textbf{82.6} & -- \\

One-template contrast
& Affirmative minus counterfactual
& -- & 64.1 \\

\bottomrule
\end{tabular}
\end{table}

\subsection{VALSE-Existence}
\label{app:valse}

Table~\ref{tab:valse} lists accuracy by evidence backend and adapter, and Table~\ref{tab:valse-negref-uncertainty} provides paired bootstrap differences against matched native CLIP.

\begin{table}[!h]
\centering
\caption{Accuracy gains over matched native CLIP (percentage points), with paired bootstrap 95\% confidence intervals. Each benchmark uses 10,000 resamples: 505 distinct images for VALSE-Existence and 402 image clusters for NegRefCOCOg. Green gains have intervals entirely above zero.}
\label{tab:valse-negref-uncertainty}

\begin{tabular}{@{}llrc@{}}
\toprule
\textbf{Scoring variant}
& \textbf{Decomposer}
& $\boldsymbol{\Delta}$
& \textbf{95\% CI} \\
\midrule

\rowcolor{backbonebg}
\multicolumn{4}{@{}c@{}}{\textbf{VALSE-Existence}} \\

Independent reflection (CLIP) & Scratch
& \textcolor{gaintext}{+8.32}
& $[+3.16,+13.47]$ \\

Counterpart anchoring (CLIP) & Scratch
& \textcolor{gaintext}{+5.94}
& $[+0.79,+11.09]$ \\

Counterpart anchoring (OWL-ViT) & Scratch
& \textcolor{gaintext}{+12.87}
& $[+7.72,+18.02]$ \\

\cmidrule(lr){1-4}

Independent reflection (CLIP) & BERT-tiny
& \textcolor{gaintext}{+7.72}
& $[+2.77,+12.67]$ \\

Counterpart anchoring (CLIP) & BERT-tiny
& \textcolor{gaintext}{+6.14}
& $[+0.99,+11.29]$ \\

Counterpart anchoring (OWL-ViT) & BERT-tiny
& \textcolor{gaintext}{+13.27}
& $[+8.32,+18.22]$ \\

\midrule

\rowcolor{backbonebg}
\multicolumn{4}{@{}c@{}}{\textbf{NegRefCOCOg}} \\

One-template contrast & Scratch
& $+2.73$
& $[-4.36,+9.84]$ \\

One-template contrast & BERT-tiny
& $+6.14$
& $[-0.92,+13.21]$ \\

\bottomrule
\end{tabular}
\end{table}

\paragraph{Caption-foil pairs:} On VALSE-Existence, independent reflection with CLIP reaches 77.6\% versus 69.3\% for native CLIP (gain $+8.32$, $[+3.16,+13.47]$); anchored reflection reaches 75.2\% ($+5.94$, $[+0.79,+11.09]$), and replacing CLIP similarity with OWL-ViT detector confidence reaches 82.2\% ($+12.87$, $[+7.72,+18.02]$). For clean pairs, anchored ordering is equivalent to thresholding detector confidence -- a score-preserving formulation, not a new decision rule. BERT-tiny gives 77.0\%, 75.4\%, and 82.6\%; the detector adds an adapter, inference cost, and a matched external boundary.

\paragraph{VALSE:} A small number of hybrid pairs retain native scores. Evaluation uses the 505 pairs with sufficient annotator agreement; the full 534-pair set is used only for image-overlap and cue-coverage checks. Detector pair ordering equals thresholding for clean positive and negation-only pairs. Calibration image identifiers are logged with zero overlap against VALSE-Existence, though this does not address whether OWL-ViT's own pretraining overlaps with COCO or Visual7W.

\begin{table}[!h]
\centering
\caption{Source-wise NegSplit performance (\%). S: from scratch; B: BERT-tiny (shaded columns). Counts list total examples / valid negated spans / valid denied spans. F1 is mean positional lexical-token F1. Bold marks the higher value within each S/B pair; ties are unbolded.}
\label{tab:decomposer-source}

\begin{tabular}{@{}llr>{\columncolor{variantbg}}r
                       r>{\columncolor{variantbg}}r
                       r>{\columncolor{variantbg}}r@{}}
\toprule
& & \multicolumn{2}{c}{\textbf{Structure acc.}}
    & \multicolumn{2}{c}{\textbf{Negated-span F1}}
    & \multicolumn{2}{c}{\textbf{Denied-span F1}} \\
\cmidrule(lr){3-4}
\cmidrule(lr){5-6}
\cmidrule(lr){7-8}
\textbf{Source} & \textbf{Counts}
& S & B & S & B & S & B \\
\midrule

COCO-MCQ & 60/39/39
& 98.3 & 98.3
& \textbf{96.6} & 95.5
& \textbf{74.5} & 72.8 \\

VOC2007-MCQ & 40/35/35
& 100.0 & 100.0
& \textbf{95.8} & 95.3
& \textbf{76.2} & 74.3 \\

\addlinespace[3pt]

COCO retrieval & 60/60/60
& 96.7 & \textbf{100.0}
& 95.2 & \textbf{97.9}
& 67.5 & \textbf{70.3} \\

MSR-VTT retrieval & 60/60/60
& 93.3 & \textbf{96.7}
& 90.8 & \textbf{91.9}
& 61.3 & \textbf{64.2} \\

\addlinespace[3pt]

VALSE & 60/33/33
& 88.3 & 88.3
& 74.1 & 74.1
& 77.1 & 77.1 \\

NegRefCOCOg & 120/117/117
& \textbf{95.8} & 95.0
& 88.5 & \textbf{92.4}
& 74.8 & \textbf{78.2} \\

\bottomrule
\end{tabular}
\end{table}

\subsection{Decomposer evaluation}
\label{app:decomposer}

Table~\ref{tab:decomposer-quality} compares NegSplit with lexical and learned baselines, Table~\ref{tab:decomposer-source} provides source-wise results, and Table~\ref{tab:downstream-spans} examines the effect of decomposition quality on downstream retrieval.

\paragraph{Annotation:} We evaluate extraction on 400 captions and expressions from the benchmark text: 60 COCO-MCQ, 40 VOC-MCQ, 60 COCO retrieval, 60 MSR-VTT retrieval, 60 VALSE, and 120 NegRefCOCOg. Two annotators each labeled a disjoint set of 200 examples from scratch, using written guidelines and text alone. Labels comprise the structural route, the negated span, and the denied-object span, which may include a denied action or scope. The set contains 56 positive, 299 hybrid, and 45 negation-only examples. Because the sets are disjoint, we report no inter-annotator agreement; metrics reflect single-annotator labels, not an agreement-validated gold standard. The same caveat applies to NegRefCOCOg's semantic-class annotations.

\paragraph{Metrics:} Route accuracy and three-class macro-F1 use all 400 examples. Positional extraction metrics compare lexical-token indices in the original caption, excluding punctuation; an exact match requires identical index sets, and F1 is averaged over examples. We locate gold labels via case-insensitive word-boundary substring matching, and all 400 examples are locatable, leaving 344 examples per extraction target. Missing or ambiguous predictions get zero credit. Forty negation-only rows have literal annotated spans shorter than the whole-caption convention used by the original scorer, and we use the literal annotations. The lexical baseline uses fixed negation cues, clause boundaries, and scaffolding-word removal without tuning on these errors. The DeBERTa~\citep{he2021deberta}-plus-rules cascade combines learned scope extraction with lexical overrides and a remainder-length route rule, and it produces no denied-object span.

\begin{table}[!h]
\centering
\caption{Downstream retrieval recall (\%) under four decomposition variants on 120 annotated queries (60 per dataset). All variants use the same captions, cached gallery, and two-template rank fusion. Bold marks the best recall per column.}
\label{tab:downstream-spans}

\begin{tabular}{@{}lrrrr@{}}
\toprule
& \multicolumn{2}{c}{\textbf{COCO}}
& \multicolumn{2}{c}{\textbf{MSR-VTT}} \\
\cmidrule(lr){2-3}
\cmidrule(lr){4-5}
\textbf{Decomposition}
& R-Neg@1 & R-Neg@5
& R-Neg@1 & R-Neg@5 \\
\midrule

Lexical parser
& 30.0 & 58.3
& \textbf{35.0} & 58.3 \\

Gold span
& 30.0 & 58.3
& 33.3 & 56.7 \\

\midrule

NegSplit (scratch)
& 30.0 & 58.3
& 33.3 & 56.7 \\

NegSplit (BERT-tiny)
& 30.0 & 58.3
& 31.7 & \textbf{60.0} \\

\bottomrule
\end{tabular}
\end{table}

\paragraph{Downstream retrieval under gold and lexical spans:} Table~\ref{tab:downstream-spans} re-ranks the 120 retrieval-sourced annotated examples under three decompositions of the same caption: NegSplit's predicted route and remainder, the human-annotated gold span, and the lexical parser's span, all with the same two-template rank fusion. On COCO, R-Neg@1 and R-Neg@5 are identical across all three (30.0 and 58.3) for both checkpoints. On MSR-VTT, NegSplit (scratch) and gold both reach 33.3 and 56.7, the lexical parser reaches 35.0 and 58.3, and BERT-tiny reaches 31.7 and 60.0. With only 60 examples per dataset, none of these differences are distinguishable from chance. The result is consistent with extraction not systematically degrading retrieval relative to gold decomposition on this subset, but it does not establish equivalence.

\subsection{Inference cost model}
\label{app:cost}

Every caption incurs one NegSplit pass, and image features and reusable text prompts can be cached. A question-level MCQ cost is
\begin{equation}
 T_Q=T_{\mathrm{img}}+\sum_{j=1}^K(T_{\mathrm{dec},j}+T_{\mathrm{native},j})+T_{\mathrm{concept}}(U)+T_{\mathrm{reference}}(Q,U),
 \label{eq:cost}
\end{equation}
where $T_{\mathrm{img}}$ covers the whole image and its $2\times2$ and $3\times3$ crops, $T_{\mathrm{dec},j}$ and $T_{\mathrm{native},j}$ are NegSplit decomposition and native affirmative-matching time for candidate $j$, $T_{\mathrm{concept}}$ covers uncached object-text encodings and crop-object similarities, and $T_{\mathrm{reference}}(Q,U)$ is the cost of computing the reflection or candidate-conditioned reference over the question's resolved categories $U$. The 14 fixed crops and the horizontal-flip view are stacked into a single batch of 15 images and encoded with one forward pass of the frozen image encoder, not 14 sequential calls. Default MCQ uses no detector pass. Retrieval reuses gallery embeddings and pays for text decomposition, affirmative text views, top-50 hybrid reranking, and full-gallery rank fusion. Table~\ref{tab:latency} excludes model initialization, media decoding, gallery construction, and one-time fold-gate moment preparation, as well as detector-backed VALSE inference. Table~\ref{tab:latency-breakdown} gives a per-component breakdown of the MCQ cost.

\begin{table}[!h]
\centering
\caption{MCQ latency breakdown under the protocol of Table~\ref{tab:latency}. Shares are percentages of the six-component sum; shading marks the dominant component. Fine-grained synchronization increases component sums by approximately 7--9\% relative to aggregate timing (Appendix~\ref{app:cost}).}
\label{tab:latency-breakdown}

\begin{tabular}{@{}lrrrr@{}}
\toprule
& \multicolumn{2}{c}{\textbf{COCO-MCQ}}
& \multicolumn{2}{c}{\textbf{VOC2007-MCQ}} \\
\cmidrule(lr){2-3}
\cmidrule(lr){4-5}
\textbf{Component}
& Time (ms) & Share
& Time (ms) & Share \\
\midrule

Crop + flip image encoding
& \cellcolor{costbg}27.5
& \cellcolor{costbg}61.2\%
& \cellcolor{costbg}24.7
& \cellcolor{costbg}58.5\% \\

Affirmative text encoding
& 7.7 & 17.1\%
& 7.7 & 18.2\% \\

Hybrid encoding + correction
& 7.1 & 15.8\%
& 7.2 & 17.1\% \\

NegSplit decomposition
& 1.4 & 3.1\%
& 1.4 & 3.3\% \\

Crop-grounded reflection scoring
& 1.1 & 2.4\%
& 1.1 & 2.6\% \\

Rank decision
& 0.1 & 0.2\%
& 0.1 & 0.2\% \\

\bottomrule
\end{tabular}
\end{table}

\paragraph{Latency breakdown:} Table~\ref{tab:latency-breakdown} decomposes batch-one MCQ latency into six components, measured on the same 200-question COCO and VOC samples, seed, and GPU as Table~\ref{tab:latency}. Crop and flip image encoding dominate, at 59--61\% of the total; the free-text hybrid denial encoding and the 2-view affirmative-remainder text encoding together account for a further 33--35\%. NegSplit parsing, crop-grounded reflection scoring, and the final rank decision together account for under 6\% of the cost: the negation-specific machinery itself is cheap, and the latency is dominated by ordinary CLIP forward passes. Instrumenting six components separately adds GPU synchronization points not present when timing the pipeline as one block, so the component sums (44.9ms COCO, 42.2ms VOC) run 7--9\% above Table~\ref{tab:latency}'s single-block aggregate (41.85ms, 38.88ms); this is measurement overhead from finer timing granularity, not a different pipeline.

\section{Robustness and Additional Analyses}
\label{app:qualify}

We provide additional analyses of candidate-pool sensitivity, the contribution of hybrid correction, performance across semantic types of negation, and coverage of the synthetic training cues relative to the benchmarks. These experiments clarify the conditions under which the reported gains hold and the limits of their interpretation.

\definecolor{stressdrop}{HTML}{F8E4E4}

\begin{table}[!h]
\centering
\caption{Accuracy (\%) with expanded candidate pools formed by borrowing distractors. CLIP uses full-caption cosine similarity; NAES uses candidate-conditioned scoring with fixed $\lambda=0.3$. Both methods use identical pools and borrowing seeds. Results average three seeds. $\Delta$ reports percentage-point changes from $K=4$; red cells highlight declines. Bold marks the higher accuracy within each matched comparison.}
\label{tab:borrowed-k}

\small

\begin{tabular}{@{}ll*{5}{r}@{}}
\toprule
& & \multicolumn{3}{c}{\textbf{Accuracy (\%)}}
& \multicolumn{2}{c}{\textbf{Change from $K=4$}} \\
\cmidrule(lr){3-5}
\cmidrule(lr){6-7}
\textbf{Dataset} & \textbf{Method}
& $K=4$ & $K=6$ & $K=8$
& $\Delta_{6}$ & $\Delta_{8}$ \\
\midrule

\rowcolor{backbonebg}
\multicolumn{7}{@{}c@{}}{\textbf{LAION-2B CLIP}} \\

\multirow{2}{*}{COCO-MCQ}
& CLIP
& 24.79 & 20.84 & 18.52
& $-3.95$ & $-6.27$ \\

& NAES
& \textbf{82.42} & \textbf{72.61} & \textbf{68.24}
& \cellcolor{stressdrop}$-9.81$
& \cellcolor{stressdrop}$-14.18$ \\

\cmidrule(lr){1-7}

\multirow{2}{*}{VOC2007-MCQ}
& CLIP
& 27.63 & 24.45 & 22.26
& $-3.18$ & $-5.37$ \\

& NAES
& \textbf{93.02} & \textbf{83.31} & \textbf{77.16}
& \cellcolor{stressdrop}$-9.71$
& \cellcolor{stressdrop}$-15.86$ \\

\midrule

\rowcolor{backbonebg}
\multicolumn{7}{@{}c@{}}{\textbf{OpenAI CLIP}} \\

\multirow{2}{*}{COCO-MCQ}
& CLIP
& 39.30 & 34.73 & 31.97
& $-4.57$ & $-7.33$ \\

& NAES
& \textbf{78.69} & \textbf{67.49} & \textbf{62.70}
& \cellcolor{stressdrop}$-11.20$
& \cellcolor{stressdrop}$-15.99$ \\

\cmidrule(lr){1-7}

\multirow{2}{*}{VOC2007-MCQ}
& CLIP
& 38.72 & 35.65 & 33.43
& $-3.07$ & $-5.29$ \\

& NAES
& \textbf{91.48} & \textbf{82.29} & \textbf{76.08}
& \cellcolor{stressdrop}$-9.19$
& \cellcolor{stressdrop}$-15.40$ \\

\bottomrule
\end{tabular}
\end{table}

\subsection{Candidate-pool stress tests}
\label{app:pool}

\paragraph{Contribution of candidate context:} The 82.4\% and 93.0\% in Table~\ref{tab:mcq} use candidate-conditioned scoring, which constructs a reference from all options in each question. Independent scoring, which scores each option independently, reaches 74.6\% and 81.4\% on the same backbone. The additional MCQ gain therefore depends partly on candidate context.

\paragraph{Accuracy under larger candidate pools:} To evaluate larger candidate pools, we append distractors borrowed from other questions, retaining only those whose false assertions are supported by target-image annotations or inherited absence facts. Table~\ref{tab:borrowed-k} shows that our accuracy decreases by approximately 9--11 percentage points with six options and 14--16 points at eight. We also evaluate native CLIP on the same candidate pools and borrowing seeds. Its accuracy declines from 24.8--39.3\% with four options to 18.5--33.4\% with eight, confirming that increasing the number of choices affects both methods. Although our framework maintains substantially higher accuracy at every pool size, its absolute decline is larger than that of native CLIP's. The hybrid correction has only a modest effect on these results. Because adding distractors changes task difficulty, candidate composition, and candidate-conditioned normalization simultaneously, this experiment cannot isolate their individual contributions. We therefore treat these results as a constructed robustness test rather than an expanded benchmark.

\begin{table}[!h]
\centering
\caption{Constructed false-distractor stress tests with OpenAI CLIP, the from-scratch decomposer, and fixed $\lambda=0.3$. Before/after are accuracies (\%); $\Delta$ is their change in percentage points, with image-cluster bootstrap 95\% confidence intervals. $n$ counts eligible questions. Red cells highlight declines; green text marks increases. These are diagnostic results, not standard benchmark scores.}
\label{tab:pool-stress}

\begin{tabular}{@{}llrrrrc@{}}
\toprule
\textbf{Distractor}
& \textbf{Scoring}
& $\boldsymbol{n}$
& \textbf{Before}
& \textbf{After}
& $\boldsymbol{\Delta}$
& \textbf{95\% CI} \\
\midrule

\rowcolor{backbonebg}
\multicolumn{7}{@{}c@{}}{\textbf{COCO-MCQ}} \\

Negation-only & Conditioned
& 5,914 & 78.69 & 82.91
& \textcolor{gaintext}{+4.22}
& $[+3.67,+4.77]$ \\

Hybrid & Conditioned
& 5,914 & 78.69 & 60.02
& \cellcolor{stressdrop}$-18.67$
& $[-19.65,-17.68]$ \\

Hybrid & Independent
& 5,914 & 70.30 & 60.18
& \cellcolor{stressdrop}$-10.12$
& $[-10.98,-9.28]$ \\

\midrule
\rowcolor{backbonebg}
\multicolumn{7}{@{}c@{}}{\textbf{VOC2007-MCQ}} \\

Negation-only & Conditioned
& 4,535 & 91.10 & 91.96
& \textcolor{gaintext}{+0.86}
& $[+0.51,+1.23]$ \\

Hybrid & Conditioned
& 2,191 & 86.26 & 75.22
& \cellcolor{stressdrop}$-11.04$
& $[-12.31,-9.80]$ \\

Hybrid & Independent
& 2,191 & 78.84 & 73.35
& \cellcolor{stressdrop}$-5.49$
& $[-6.57,-4.42]$ \\

\bottomrule
\end{tabular}
\end{table}

\paragraph{Borrowed-option protocol:} Borrowed options are admitted only when their false assertion is supported by target-image annotations or a lack of an inherited benchmark fact. We reran candidate-conditioned NAES scoring and the fixed-$0.3$ correction on the target image. Results average three fixed borrowing seeds and exact option permutations for K4 and K6, plus 512 balanced orders for K8.

\begin{wraptable}{r}{0.46\linewidth}
\centering
\setlength{\tabcolsep}{3pt}
\renewcommand{\arraystretch}{1.10}
\caption{Effect of hybrid correction ($\lambda=0.3$) with LAION-2B and the from-scratch decomposer. $\Delta$ is corrected minus uncorrected accuracy (percentage points), with paired image-cluster 95\% confidence intervals. Results use complete datasets and average over 24 option orders. Green cells mark gains whose intervals exclude zero.}
\label{tab:mcq-uncertainty}
{\scriptsize
\begin{tabular}{@{}llrc@{}}
\toprule
\textbf{Dataset} & \textbf{Scoring}
& $\boldsymbol{\Delta}$ & \textbf{95\% CI} \\
\midrule
COCO & Independent
& \cellcolor{cigainbg}$+0.56$
& $[+0.28,+0.85]$ \\
& Conditioned
& $+0.05$
& $[-0.22,+0.33]$ \\
\midrule
VOC2007 & Independent
& \cellcolor{cigainbg}$+1.62$
& $[+1.22,+2.03]$ \\
& Conditioned
& $-0.02$
& $[-0.29,+0.26]$ \\
\bottomrule
\end{tabular}
\par}
\end{wraptable}

\paragraph{Misleading hybrid distractors:} In the false-distractor test (Table~\ref{tab:pool-stress}), we replace a wrong option with a constructed false negation. For negation-only options, the denied object is confirmed present; for hybrid options, both the affirmed and denied objects are confirmed present. Accuracy rises with the inserted negation-only distractors but falls sharply with hybrid distractors, by 18.67 points on COCO with candidate-conditioned scoring and 10.12 with pool-independent scoring. The inserted option is false, yet the scoring procedure can favor it; in the candidate-conditioned route, its presence can also change scores assigned to the other options. This exposes a limitation of denied-object evidence and relative candidate scoring rather than verifying the negative claim.

\paragraph{False-distractor protocol:} The new text is processed by the fixed NegSplit checkpoint and fixed-$\lambda$ scoring pipeline. We retain the correct option text, but recomputing the candidate-conditioned pool can change its score; the correct option's score is not guaranteed to remain unchanged.

\subsection{The hybrid correction}
\label{app:correction}

On LAION-2B, the fixed-$\lambda$ correction changes candidate-conditioned MCQ accuracy by $+0.05$ (COCO) and $-0.02$ (VOC) points, with intervals including zero; it contributes more under independent scoring, $+0.56$ and $+1.62$ (Table~\ref{tab:mcq-uncertainty}). In retrieval, gains are clearest on OpenAI COCO, smaller on LAION-2B COCO R-Neg@1, and uncertain on MSR-VTT and LAION-2B R-Neg@5 (Table~\ref{tab:retrieval-correction}). The correction is a secondary, setting-dependent refinement.

\begin{table}[tbh]
\centering\small
\caption{R-Neg recall (\%) before and after the fixed $\lambda=0.3$ hybrid correction, with paired image or video-cluster 95\% confidence intervals for the change. The baseline is affirmative two-view rank fusion, and corrected scores are reranked within the same top-50 shortlist before full-gallery rank fusion.}
\label{tab:retrieval-correction}
\resizebox{\linewidth}{!}{%
\begin{tabular}{@{}llrrrrrr@{}}
\toprule
Backbone & Dataset & R-Neg@1 base & Corrected & $\Delta$ (95\% CI) & R-Neg@5 base & Corrected & $\Delta$ (95\% CI) \\
\midrule
LAION-2B & COCO & 39.52 & 39.93 & $+0.41$ [$+0.12,+0.71$] & 64.44 & 64.68 & $+0.24$ [$-0.01,+0.50$] \\
 & MSR-VTT & 31.30 & 31.90 & $+0.60$ [$-0.70,+1.90$] & 54.20 & 54.00 & $-0.20$ [$-1.50,+1.10$] \\
OpenAI & COCO & 31.74 & 33.27 & $+1.53$ [$+1.24,+1.80$] & 54.89 & 56.40 & $+1.51$ [$+1.23,+1.79$] \\
 & MSR-VTT & 30.40 & 30.60 & $+0.20$ [$-1.10,+1.50$] & 53.30 & 54.00 & $+0.70$ [$-0.70,+2.10$] \\
\bottomrule
\end{tabular}}
\end{table}

\subsection{NegRefCOCOg}
\label{app:negref}

Table~\ref{tab:negref-per-class} gives the locally annotated semantic composition of the 440 expressions.

\begin{wraptable}{r}{0.45\linewidth}
\centering
\vspace{-2em}
\caption{NegRefCOCOg composition and accuracy (\%). Shares use all 440 expressions; bold marks the best accuracy per row. Rules denotes the local clause parser; NS denotes NegSplit's one-template contrast.}
\label{tab:negref-per-class}

\scriptsize

\begin{tabular}{@{}lrrrrr@{}}
\toprule
& \multicolumn{2}{c}{\textbf{Composition}}
& \multicolumn{3}{c}{\textbf{Accuracy}} \\
\cmidrule(lr){2-3}
\cmidrule(lr){4-6}
\textbf{Class}
& $n$ & Share
& CLIP & Rules & NS \\
\midrule

Existence
& 246 & 55.9
& 61.0 & \textbf{67.9} & 62.6 \\

Attribute/state
& 58 & 13.2
& 56.9 & \textbf{65.5} & 44.8 \\

Event
& 83 & 18.9
& 51.8 & \textbf{72.3} & 61.5 \\

Relation
& 22 & 5.0 & 45.5 & 59.1 & \textbf{63.6} \\

Others
& 31 & 7.0 & 61.3 & 58.1 & \textbf{71.0} \\

\midrule
Total
& 440 & 100.0
& 58.0 & 67.3 & 60.7 \\
\bottomrule
\end{tabular}
\end{wraptable}

\paragraph{Referring-expression contrast:} NegRefCOCOg presents two candidate regions of the same object category and asks which one the expression describes. Because the expressions are short and the negated clause is lexically marked, we reuse NegSplit's decomposition interface directly instead of the three-way dispatch. This is a separate application of the decomposer, not a fourth scoring route. The affirmative counterfactual $\tilde b(t)$ is formed by a fixed token-level polarity rewrite that flips the negated clause to affirmative in place (``no shirt'' to ``a shirt,'' ``not visible'' to ``visible'') and reattaches it to the affirmed part; for example, ``silver refrigerator'' plus ``is not open'' gives ``silver refrigerator that is open.'' The rewrite is applied uniformly across existence, attribute, event, and relation negation. The contrast score compares two text views against the image,
\begin{equation}
  S_{\mathrm{contrast}}(I,t)=s_{\mathrm{VLM}}\big(I,a_{\mathrm{raw}}(t)\big)-s_{\mathrm{VLM}}\big(I,\tilde b(t)\big),
  \label{eq:contrast}
\end{equation}
where $a_{\mathrm{raw}}(t)$ is the raw affirmative remainder, which keeps leftover connecting words such as a dangling ``but,'' and both views drop terminal punctuation. A single unwrapped text view is used, not the two-template fusion, because the task is a direct two-way comparison. Contrast requires a predicted hybrid label, a nonempty raw remainder, and a nonempty counterfactual; otherwise, the original phrase is scored. Both regions belong to the same category, so similarity to that category cannot separate them; instead, contrast measures whether the image better matches the literal claim or its affirmed counterfactual. For attribute, event, and relation negation, the denied proposition is not an entity, so object-presence evidence does not address it, which is why we score the claim as text. The rewrite is deterministic and can introduce errors; the NegSplit-based contrast trails the dataset-specific rule-based reference (67.3\%).

\paragraph{Negation beyond object existence:} NegRefCOCOg includes all four semantic classes but contains only 440 expressions. The one-template affirmative-minus-counterfactual contrast reaches 60.7\% (scratch) and 64.1\% (BERT-tiny), compared with 58.0\% for matched CLIP. Both paired intervals for the gains ($+2.73$ with $[-4.36,+9.84]$ and $+6.14$ with $[-0.92,+13.21]$) include zero; at this sample size, neither overall improvement is statistically distinguishable from zero. Table~\ref{tab:negref-per-class} shows uneven performance: the scratch contrast scores 62.6\% versus 61.0\% on existence negation, but 44.8\% versus 56.9\% on attribute/state negation. Numerical improvements for event and relation negation are exploratory because the class counts are small. These findings do not establish broad semantic generalization.

\subsection{Synthetic-cue coverage}
\label{app:cue}

\begin{wraptable}{r}{0.5\linewidth}
\centering
\caption{Coverage by NegSplit's training cue vocabulary. A string is covered if a cue matches case-insensitively at word boundaries. This measures literal cue overlap, not training-example overlap.}
\label{tab:template-overlap}

\scriptsize
\setlength{\tabcolsep}{3pt}
\renewcommand{\arraystretch}{1.15}

\begin{tabular}{@{}lrrr@{}}
\toprule
\textbf{Benchmark}
& \textbf{Scanned}
& \textbf{Covered}
& \textbf{Coverage} \\
\midrule

COCO-MCQ
& 23,656 & 15,503
& \cellcolor{coveragebg}65.5\% \\

VOC2007-MCQ
& 20,124 & 14,241
& \cellcolor{coveragebg}70.8\% \\

VALSE-Existence
& 1,068 & 534
& \cellcolor{coveragebg}50.0\% \\

NegRefCOCOg
& 440 & 440
& \cellcolor{coveragebg}100.0\% \\

\bottomrule
\end{tabular}

\par\smallskip
{\scriptsize\raggedright
Scanned strings include all MCQ options, both captions and foils for VALSE, and all NegRefCOCOg phrases.\par}
\end{wraptable}

Table~\ref{tab:template-overlap} scans every COCO-MCQ and VOC2007-MCQ option, VALSE caption and foil, and NegRefCOCOg phrase for a case-insensitive word-boundary match against the literal negation-cue strings used to build the 181,734 synthetic training examples. At least one cue matches 65.5\% of COCO-MCQ options, 70.8\% of VOC2007-MCQ options, 50.0\% of VALSE strings (exactly one member of each pair is negated, as expected), and 100\% of NegRefCOCOg phrases. No string containing an unambiguous negation signal (``no,'' ``not,'' ``n't,'' or ``without'') was left unmatched. This shows the synthetic cue vocabulary is not narrower than the benchmarks' negation phrasing, but it does not establish that NegSplit's predictions are not pattern-matched to surface cue forms.

\end{document}